%% file: main.tex
\documentclass[10pt,twocolumn,letterpaper]{article}

\usepackage[pagenumbers]{cvpr} 

\usepackage{graphicx}
\usepackage{amsmath}
\usepackage{amssymb}
\usepackage{booktabs, comment}

\usepackage{url}
\usepackage[utf8]{inputenc} 
\usepackage[T1]{fontenc}    
\usepackage[table,xcdraw,dvipsnames]{xcolor}
\usepackage{subcaption}
\usepackage[normalem]{ulem} 
\usepackage[export]{adjustbox}
\usepackage{amsfonts}       
\usepackage{nicefrac}       
\usepackage{microtype}      
\usepackage{gensymb}
\usepackage{tabularx}
\usepackage{siunitx}
\usepackage{enumitem}
\usepackage{lineno}
\usepackage{pifont}
\usepackage[square,numbers]{natbib}
\usepackage{upgreek}
\usepackage{bbm}

\usepackage[dvipsnames]{xcolor}

\usepackage[font=small,labelfont=bf]{caption}

\newcommand{\RN}[1]{%
  \textup{\uppercase\expandafter{\romannumeral#1}}%
}

\definecolor{navy}{rgb}{0.7, 0.1, 0.7}
\newcommand{\KGcr}[1]{{\color{black}{}#1}} 

\definecolor{burgundy}{RGB}{144,0,32}
\newcommand{\SM}[1]{{\color{black}#1}} 
\newcommand{\SMC}[1]{{\color{black}#1}} 

\usepackage[pagebackref,breaklinks,colorlinks]{hyperref}

\usepackage[capitalize]{cleveref}
\crefname{section}{Sec.}{Secs.}
\Crefname{section}{Section}{Sections}
\Crefname{table}{Table}{Tables}
\crefname{table}{Tab.}{Tabs.}

\def\cvprPaperID{11582} 
\def\confName{CVPR}
\def\confYear{2023}

\usepackage{fancyhdr}

\fancypagestyle{firstpage}{%
  \fancyhf{}
  \chead{\small In Proceedings of IEEE/CVF Conference on Computer Vision and Pattern Recognition (CVPR), 2023}
}

\begin{document}

\title{Chat2Map: Efficient Scene Mapping from Multi-Ego Conversations}

\author{
Sagnik Majumder$^{1, 2, 3}$ \hspace{3mm} Hao Jiang$^{2}$  \hspace{3mm} Pierre Moulon$^{2}$ \hspace{3mm}  Ethan Henderson$^{2}$\\ Paul Calamia$^{2}$  \hspace{3mm} Kristen Grauman$^{1,3}$\thanks{Equal contribution} \hspace{3mm} Vamsi Krishna Ithapu$^{2*}$\\
$^1$UT Austin \hspace{3mm} $^2$Reality Labs Research, Meta \hspace{3mm} $^3$FAIR
}
\maketitle

\thispagestyle{firstpage}

\input{sections/abstract}

\input{sections/introduction}

\input{sections/related_work}

\input{sections/task}

\input{sections/approach}

\input{sections/experiments}

\input{sections/conclusion}

{\small
\bibliographystyle{ieee_fullname}
\bibliography{mybib}
}

\clearpage

\input{sections/supp}

\end{document}

%% file: sections/abstract.tex
\begin{abstract}

Can conversational videos captured from multiple egocentric viewpoints reveal the map of a scene in a cost-efficient way?  We 
 seek to answer this question by proposing a new problem: efficiently building  the map of a previously unseen 3D environment by exploiting  shared information in the egocentric audio-visual observations of participants in a natural conversation. Our hypothesis is that as multiple people (``egos") move in a scene and talk among themselves, they receive rich audio-visual cues that can help uncover the unseen areas of the scene.  
Given the high cost of continuously processing egocentric visual streams, we further explore how to actively coordinate the sampling of visual information, so as to minimize redundancy and reduce power use. To that end, we present an audio-visual deep reinforcement learning approach that works with our shared scene mapper to selectively turn on the camera to efficiently chart out the space. 
We evaluate the approach using a state-of-the-art audio-visual simulator for 3D scenes as well as real-world video.  Our model outperforms previous state-of-the-art mapping methods, and achieves an excellent cost-accuracy tradeoff. Project: \url{http://vision.cs.utexas.edu/projects/chat2map}.
\end{abstract}

%% file: sections/introduction.tex
\vspace{-0.25cm}
\section{Introduction}\label{sec:intro}

\begin{figure}[t] 
    \centering
    \includegraphics[width=1\linewidth]{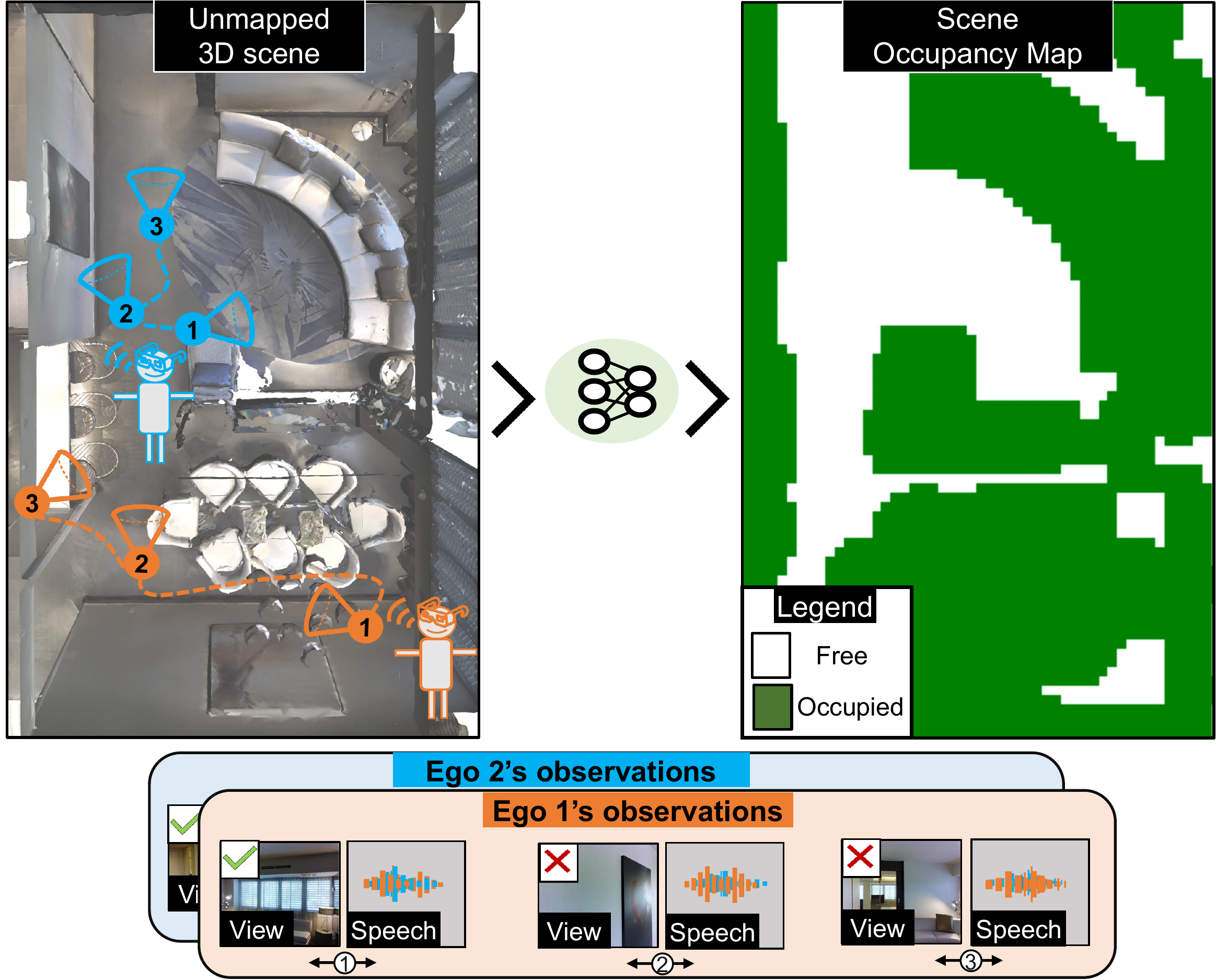}
    \caption{
    Given egocentric audio-visual observations from multiple people wearing AR glasses and moving and conversing  (left), we aim to accurately map the scene (right).
    To mitigate cost, our model receives audio continuously but learns to selectively employ the ego cameras only when the visual input is expected to be informative.
    }
    \vspace*{-0.12in}
\label{fig:intro}
\end{figure}

The spatial layout of the environment around us is fundamental to understanding our physical context. 
By representing the walls, furniture, and other major structures in a space, \emph{scene maps} ground activity and objects in a persistent frame of reference, facilitating high-level reasoning for many downstream applications in augmented reality (AR) and robotics.  
For example, episodic memory~\cite{ego4d,datta-em} aims to relocalize lost objects observed in first-person  video (\emph{where are my keys?}); floorplan estimation~\cite{liu2018floornet,chen2019floor,Okorn2010TowardAM} aims to chart out the area and shapes of complex buildings; navigating agents try to discover routes in unfamiliar spaces~\cite{Chaplot2020Learning,chen2019learning,ramakrishnan2020occupancy}.

While traditional computer vision approaches for mapping (e.g., visual SLAM) are highly effective when extensive exposure to the environment is possible, in many real-world scenarios only a fraction of the space is observed by the camera.  Recent work shows the promise of sensing 3D spaces with both sight and sound~\cite{purushwalkam2021audio,batvision,10.1007/978-3-030-58545-7_38,tdw,chen2020soundspaces}: listening to echoes bounce around the room can reveal the depth and shape of surrounding surfaces, and even help extrapolate a floorplan beyond the camera's field of view or behind occluded objects~\cite{purushwalkam2021audio}.  

While we are inspired by these advances, they also have certain limitations.  Often systems will emit sounds (e.g., a frequency sweep) into the environment to ping for spatial information~\cite{batvision,10.1007/978-3-030-58545-7_38,purushwalkam2021audio,8953208,7012073,Eliakim2018AFA, 10.1162/isal_a_00222, 9196934}, which is intrusive if done around people. Furthermore, existing audio-visual models assume that the camera is \emph{always on} grabbing new frames, which is wasteful if not intractable, particularly on lightweight, low-power computing devices in AR settings.

We introduce Chat2Map, a new scene mapping task aimed at eliminating these challenges.  
In the proposed setting, multiple people converse as they move casually through the scene while wearing AR glasses equipped with an egocentric camera, microphones, and potentially other sensors (e.g., for odometry).\footnote{Throughout, we call each person participating in the conversation an ``ego" for short.}  Given their egocentric audio-visual data streams, the goal is to infer the ground-plane occupancy map for the larger environment around them.  See Figure~\ref{fig:intro}.

We observe that audio-visual data from the egos' interactions will naturally reflect scene structure.  First, as they walk and talk, their movements reveal spaces like corridors, doorways, and large rooms, in both modalities.  Second, the speech captured by the device-wearer's cameras and microphones can be localized to different speakers, which, compared to active sound emission, is non-intrusive. 
 
To realize this vision, we develop a novel approach to efficient scene mapping from multi-ego conversations.  Our approach has two key elements: a shared scene mapper and a visual sampling policy.
For the former, we devise a transformer-based mapper that incorporates the multiple data streams to infer a  map beyond the directly observed areas, and, most importantly, that enables communication among the egos about their observations and states in the 3D space to improve mapping accuracy.  For the latter, our idea is to relax the common assumption of an ``always-on" camera, and instead \emph{actively select} when to sample visual frames from any one of the ego cameras.  Intuitively, certain regions where the egos move will be more or less important for mapping (e.g., corners of the room, doors).  We train a sampling policy with deep reinforcement learning that activates the visual feed only when it is anticipated to complement the continuous audio feed.  This is a cost-conscious approach, mindful that 
switching on a camera is much more power consuming than sensing audio with microphones
\cite{carroll2010analysis, likamwa2014draining}.

We demonstrate our approach using a state-of-the-art audio-visual simulator for 3D scenes~\cite{chen2020soundspaces} as well as real-world video input.  We can successfully map an unfamiliar environment given only partial visibility via multiple conversing people moving about the scene. 
Compared to sampling all visual frames, our model reduces the visual \SMC{capture and} processing by $87.5\%$ while the mapping accuracy declines \KGcr{only} marginally ($\sim 9\%$).

\SMC{Our main contributions are 1) we define the task of efficient and shared scene mapping from multi-ego conversations, a new direction for AR/VR research; 2) we present the first approach to begin tackling this task, specifically an RL-based framework that integrates an audio-visual scene mapper and a smart visual sampling policy; and 3) we rigorously experiment with a variety of environments (both in simulation and the real world), speech sounds, and use cases.}

%% file: sections/related_work.tex
\section{Related Work}\label{sec:related}

\paragraph{Visual scene mapping.} 
Past works tackle scene mapping using 3D Manhattan layouts~\citep{zou2018layoutnet, yang2019dula, sun2019horizonnet, dhamo2019object, zou2021manhattan}, detailed floorplans~\citep{Okorn2010TowardAM, 7346513, liu2018floornet, chen2019floor, 10.1145/3355089.3356556}, occupancy~\citep{10.1177/0278364911421039, doi:10.1177/0278364916684382, Senanayake2017DOM, 8793500, 8793769, elhafsi2020map}, and semantic maps~\citep{narasimhan2020seeing}. Manhattan layouts include structured outputs like scene boundaries~\citep{zou2018layoutnet, sun2019horizonnet, zou2021manhattan}, corners~\citep{zou2018layoutnet, zou2021manhattan}, and floor/ceilings~\citep{yang2019dula, zou2021manhattan}, but do not generalize to unseen environment regions. Floorplan estimation methods use dense scans of 3D scenes to predict geometric (walls, exterior/ interior) and semantic layouts (room type, object type, etc.), rely on extensive human walkthroughs with RGB-D~\citep{liu2018floornet, chen2019floor} or 3D point cloud~\citep{Okorn2010TowardAM, 7346513} scans, and are usually limited to polygonal layouts~\citep{Okorn2010TowardAM, 7346513, liu2018floornet, chen2019floor, 10.1145/3355089.3356556}. 
\SM{Occupancy maps} traditionally rely on wide field-of-view (FoV) LiDAR scanners~\citep{6719348} or evaluate  on simple 2D environments without non-wall obstacles~\citep{8793500, 8793769, elhafsi2020map, 8793769}. More recent methods~\citep{chen2019learning, chaplot2020object, ramakrishnan2020occupancy, Chaplot2020Learning} train an embodied agent to explore and build top-down maps of more complex scenes using RGB-D. On the contrary, our method uses both vision and audio \SM{from the observations of a group of conversing people for mapping.}  Rather than steer the camera of a robot to map the scene, our task requires processing passive video from human camera wearers.

\vspace*{-0.15in}
\paragraph{Audio-visual scene mapping.}
\SM{To our knowledge, the only prior work to translate audio-visual inputs into a general (arbitrarily shaped) floorplan maps is AV-Floorplan~\citep{purushwalkam2021audio}.  Unlike AV-Floorplan, our method \KGcr{infers} maps from speech in natural human conversations, which avoids emitting intrusive frequency sweep signals to generate echoes. In addition,
a key goal of our work is to reduce mapping cost by skipping redundant visual frames. 
Our experiments demonstrate the benefits of our model design over AV-Floorplan~\citep{purushwalkam2021audio}.}

\vspace*{-0.15in}
\paragraph{Audio(-visual) spatial understanding.}
More broadly, beyond the mapping task, various methods leverage audio for geometric and material information about the 3D scene and its constituent objects. Prior work 
relies on acoustic reflections to estimate the shape of objects~\citep{8953208}. Echolocation is used in robotics to estimate proximity to surrounding surfaces~\citep{7012073, Eliakim2018AFA, 10.1162/isal_a_00222, 9196934}. 
Together, vision and audio can better reveal the shape and materials of objects~\citep{7780633, Zhang2017GenerativeMO, Schissler2018AcousticCA}, self-supervise imagery~\citep{10.1007/978-3-030-58545-7_38}, and improve depth sensing
~\citep{7299122, 8374617}. 
Recent work exploits correlations between spatial audio and  imagery to reason about scene acoustics~\citep{Chen2022VisualAM, Majumder2022FewShotAL} or
aid active embodied navigation~\citep{dean2020see,9197008, chen2021learning, chen2021semantic, yu2022sound} and source separation~\citep{majumder2021move2hear, majumder2022active}. No prior work
intelligently \KGcr{selects images to capture} 
during conversations to efficiently map a scene.

\vspace*{-0.15in}
\paragraph{Multi-agent spatial understanding.} \SM{There is existing work~\citep{Das2019TarMACTM, Jain2019TwoBP, iqbal2019coordinated, Jain2020ACS, Patel2021InterpretationOE} in the visual multi-agent reinforcement learning (MARL) community that learns collaborative agents for performing tasks like 
relocating furniture~\citep{Jain2019TwoBP, Jain2020ACS}, playing 3D multi-player games~\citep{Jaderberg2019HumanlevelPI}, coordinated scene exploration~\citep{iqbal2019coordinated}, or multi-object navigation~\citep{Patel2021InterpretationOE}. In such settings, the collaborative agents actively interact with the environment to learn a shared scene representation
for successfully completing their task. In contrast, we aim to learn a shared geometric map of a 3D scene given \emph{passive} observations that come from the trajectories chosen by a group of people involved in a natural conversation.}

\vspace*{-0.15in}
\paragraph{Efficient visual sampling in video.} 
Efficient visual sampling has been studied in the context of video recognition~\citep{yeung2016end, Korbar2019SCSamplerSS, Gao2020ListenTL, 
lin2022ocsampler, wu2019adaframe}
and summarization~\citep{chen2018less, Suin_Rajagopalan_2020} with the goal of \KGcr{selectively} 
processing informative frames, which can both reduce computational cost and improve recognition.  
More closely related to our approach are methods that use audio for the decision-making~\citep{Korbar2019SCSamplerSS, Gao2020ListenTL, panda2021adamml}. 
Different from the above, we use efficient visual sampling in the context of mapping scenes.
Furthermore, in our case an online sampling decision needs to be made at every step \emph{before} looking at the current visual frame (or frames from future steps).

%% file: sections/task.tex
\section{
Chat2Map Task Formulation} 
\label{sec:task}

We propose a novel task: efficient and shared mapping of scenes from multi-ego 
conversations. 

\begin{figure*}[t] 
    \centering
    \includegraphics[width=.85\linewidth]{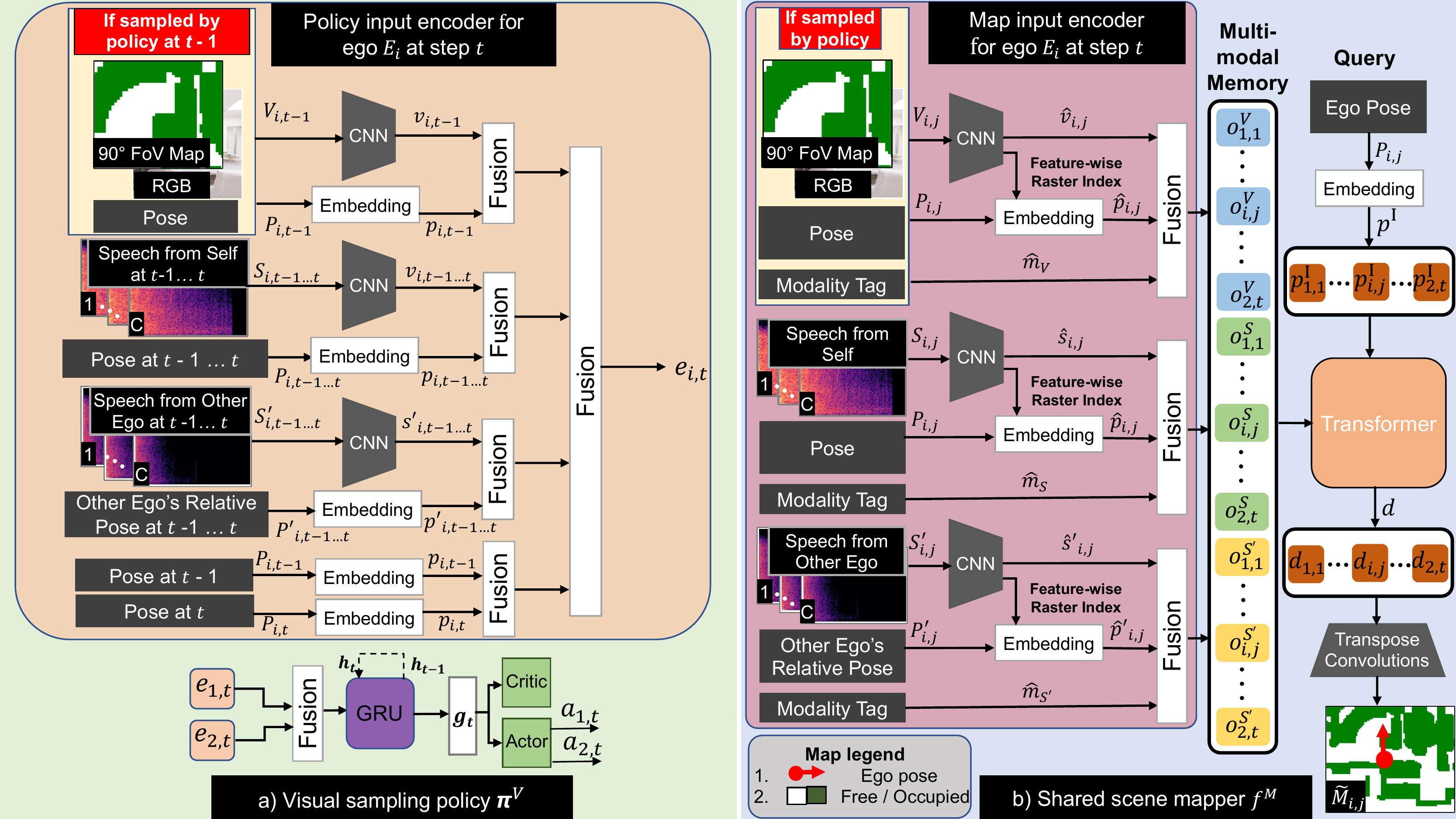}
    \vspace*{-0.1in}
    \caption{Our model has two main components: a) a visual sampling policy (left), and b) a shared scene mapper (right). At each step, our policy receives the current audio along with the previous audio(-visual) observations for the egos and decides \SM{for each ego individually} whether to capture its visual frame 
    at the current step. 
    As per the policy predictions, the shared mapper conditionally uses the current visual frame(s) and audio along with the past audio(-visual) observations to predict the occupancy map of the scene, a ground-plane map showing where obstacles and freespace are (shown in green and white).
}
\vspace*{-0.15in}
\label{fig:model}
\end{figure*}

\SM{Without loss of generality, we consider two egos, $E_1$ and $E_2$,  each}
wearing AR glasses 
equipped with an RGB-D camera and a 
 multi-channel
microphone array. 
The egos have a conversation and move around in an unmapped 3D environment.
Each conversation is $T$ steps long. 
At each step $t$,the ego $E_i$'s glasses receive an observation 
\SM{$\mathcal{O}_{i, t} = (\mathcal{V}_{i, t}, S_{i, t}, P_{i, t}, S^{'}_{i, t}, P^{'}_{i, t})$}.
$\mathcal{V}_{i, t}$ 
is the $90^\circ$ FOV RGB-D image and $\mathcal{S}_{i,t}$ is the speech waveform uttered by $E_i$, as observed from its pose $P_{i, t} = (x_{i,t}, y_{i,t}, \theta_{i,t})$, where $(x_{i,t}, y_{i,t})$ denotes its location and $\theta_{i,t}$ denotes its orientation in the 3D scene. \SM{$\mathcal{S}^{'}_{i, t}$} is the speech of the other ego $E^{'}_i$ (\SM{the other} person \SM{involved in the conversation}), as perceived by $E_i$ (note, the speech sounds different depending on the listener position), and \SM{$P^{'}_{i, t}$ is $E^{'}_i$}'s pose relative to $E_i$. \SM{Modern AR glasses, like Bose Frames or Facebook Aria already support capturing such multi-sensory observations, making it possible to have a real-world instantiation of our task.}

Given the \SM{real-time observation stream $\mathcal{O}$ for the egos, where}  $\mathcal{O} = \big\{\mathcal{O}_{i,t}: i = 1,\ldots,2, t = 1 \ldots T\big\}$ 
and a total budget of visual frames $B$,
we aim to learn a model that can \SM{accurately} estimate the top-down occupancy map $M$ of the scene \SM{without  exceeding the visual budget}.
\SMC{The fixed budget \KGcr{represents allotting} 
a fixed amount of battery to the energy-intensive visual capture~\citep{likamwa2014draining}, so that other on-device routines are not affected.}
We assume the first visual frames (at $t=1$) for both egos to be used by \SM{the model}.
Thus  we aim to learn a policy that samples $B$ frames from $2*(T-1)$ choices---which are not considered a batch, but rather unfold in sequence---\SM{and a mapper that predicts the scene map given the sampled frames}. 
\KGcr{Our} goal is to build a 
model that 
samples the expensive visual frames only when absolutely needed \SM{for scene mapping}. 
This is captured by the constraint $1 \leq B <\!<\! 2*(T-1)$. 

\SM{There are three important aspects to our task. First, it}
requires learning from both vision and audio. While the visual signal carries rich information about the local scene geometry, there can be a high amount of redundancy in the visual feed captured during a conversation (\eg, the egos may visit the same location more than once or change their viewpoint only marginally). 
\SM{Second}, not only does the long-range nature of audio help uncover the global scene properties~\citep{doi:10.1073/pnas.1221464110, purushwalkam2021audio} like shape and size---beyond what is visible in images--- 
\SM{\KGcr{but also} we can exploit audio} to undersample the visual frames,
thereby reducing the cost of capturing and processing sensory inputs for mapping. 
\SM{Third}, \KGcr{\emph{shared}} mapping of a scene \KGcr{allows}
jointly leveraging the complementary information in the audio (speech) from the self and other egos,
and the synergy of the audio-visual cues from multiple egos. 
These insights form the basis of our key hypothesis in this task---selectively sampling visual frames during a conversation involving egos that share information with each other can facilitate efficient mapping of a scene.

%% file: sections/approach.tex
\section{Approach}\label{approach}

We solve the task 
\SM{by learning a model that estimates the scene map given the egos' audio-visual observations and also}
sequentially decides when 
to sample visual frames 
\SM{for mapping} given the \SM{audio stream, ego poses,}
and previously sampled

frames, if any. \SM{Here, "sampling" refers to \emph{individually} deciding for \emph{each} ego whether to use its camera or not to capture the visuals 
at every step of its trajectory in the scene.}
\SM{The sampling is} preemptive in nature, \ie the policy selects or 
\SM{skips}
a frame \emph{without capturing it first}.

\SM{Our model} has two main components (see Fig.~\ref{fig:model}): 
\textbf{(1)} a shared scene mapper, and \textbf{(2)} a visual sampling policy. 
At every step $t$, the shared mapper has two functions. First, it estimates the map of a previously unseen environment by exploiting the shared spatial cues 
in the audio-visual observations of the two egos.
Second, it informs the policy about the utility of sampling a certain visual frame. 
Guided by the mapper, the policy samples only the most informative visual frames that \KGcr{are expected to}
boost mapping significantly over using just audio.  
Note that, unlike the visuals, we observe audio continuously \SM{as it is 
less resource-intensive vis-a-vis storage and power requirements for processing}~\citep{carroll2010analysis}. 

We learn our task through the synergy of the mapper and the policy, such that under the constraint of a limited visual budget $B$, 
our model implicitly understands 
which visual frames are critical for mapping.

First, we describe the steps involved to prepare our model inputs (Sec.~\ref{sec:inp_prep}). Next, we introduce our visual sampling policy (Sec.~\ref{sec:policy}) and shared scene mapper (Sec.~\ref{sec:mapper}). Finally, we present model training details (Sec.~\ref{sec:training}). \SM{Through the rest of the text, we use separate notations to distinguish the egos' observations $\mathcal{O}$ (\ie what the egos \emph{receive} from the \emph{environment}) from our model inputs $O$ (\ie what we \emph{capture} and \emph{feed} to our model for efficient mapping).}

\subsection{Model input preparation}\label{sec:inp_prep}

We prepare our model inputs by separately preprocessing the visual and audio modalities. 
\SM{If our policy decides to sample an image $\mathcal{V}$, we transform it} into $V = (V^R, V^M)$. 
$V^R$ denotes the normalized RGB image
with pixel values $\in [0, 1]$. 
$V^M$ denotes the $90^\circ$ FoV top-down occupancy map created by projecting the depth image \KGcr{to the ground plane}.
To do the depth projection, we 
first backproject it into the world coordinates using the camera's 
intrinsic parameters to compute the local visible scene's 3D point cloud. 
Next, we project these points to obtain a two-channel binary top-down map of size $h \times w \times 2$, 
where the first channel 
reveals occupied/free areas, and the second channel reveals seen/unseen areas. \SM{If our policy skips $\mathcal{V}$}, we set
$V^R$ and $V^M$ to all-zero matrices of the appropriate size.

For a speech waveform $\mathcal{S}$, 
we calculate the short-time Fourier transform (STFT) magnitude spectrogram 
denoted by $S$ 
of size $F \times \mathcal{T} \times C$, 
where $F$, $\mathcal{T}$, and $C$ are the number of frequency bins, time windows, and \SM{ambisonic microphone} channels, respectively. 
Lastly, we normalize each pose $P_{i, t}$ to be relative to $P_{1,1}$.
See \SM{Sec.~\ref{sec:experiments} and} Supp. for more details. 

\subsection{Visual sampling policy}\label{sec:policy}

At every step $t$, our visual sampling policy $\pi^V$ (Fig.~\ref{fig:model} left) receives $O^{\pi}(t)$ as input 
and makes the decision to either capture or \SM{skip} the visual frame $\mathcal{V}_{i, t}$ for each ego $E_i$. $O^{\pi}(t)$ comprises the visual cue from the last step along with the speech cues, 
and the poses from the current step and the last step for both egos. 
Formally, $O^\pi(t) = \big\{O^{\pi}_{i}(t): i = 1 \ldots 2\big\}$, where $O^{\pi}_{i}(t)  = \big\{V_{i, t-1}, S_{i, j}, P_{i, j}, S^{'}_{i, j}, P^{'}_{i, j}: j = t-1 \ldots t\big\}$. 
The policy first uses an encoder network to generate a multi-modal embedding of $O^{\pi}(t)$, 
and then passes the embedding to a policy network that makes a sampling decision per ego. 
At $t = 1$, as per our problem definition (Sec.~\ref{sec:task}), the policy always chooses to sample the visual frames for both egos, i.e., the cameras are initially on.   

\vspace*{-0.15in}
\paragraph{Multi-modal policy embedding.}
To process ego $E_i$'s visual input $V_{i, t-1}$ from the last step, we encode the RGB image $V^R_{i, t-1}$ and map $V^M_{i, t-1}$ with separate CNNs.
We then concatenate the two features to generate the visual embedding \SM{$v_{i, t-1}$.}
To encode the 
\SM{pose inputs}
$\big\{P_{i, t-1}, P^{'}_{i, t-1}, P_{i, t}, P^{'}_{i, t}\big\}$, 
we use a linear layer and generate pose embeddings 
\SM{$\big\{p_{i, t-1}, p^{'}_{i, t-1}, p_{i, t}, p^{'}_{i, t}\big\}$}. 
We process the 
speech inputs 
$\big\{S_{i, t-1}, S^{'}_{i, t-1}, S_{i, t}, S^{'}_{i, t}\big\}$ 
using another CNN and create speech embeddings 
\SM{$\big\{{s}_{i, t-1}, {s}^{'}_{i, t-1}, {s}_{i, t}, {s}^{'}_{i, t}\big\}$}. 
Next, we \SM{fuse the visual, speech and pose embeddings using linear layers (see Fig.~\ref{fig:model} left)}
to obtain the multi-modal policy embedding $e_{i, t}$ for $E_i$. Finally, we fuse the policy embeddings for the two egos, $e_{1, t}$ and $e_{2, t}$ with a linear layer to produce the multi-modal policy embedding $e_{t}$. 

The visual, audio, and pose inputs carry complementary cues required for efficient visual sampling.
Whereas the pose inputs from the last and current steps explicitly reveal the viewpoint change 
between the steps, the previous and current speech inputs provide  
information about the changes in the  local and global scene structures as a function of the previously sampled visual inputs, which together suggest the value of
sampling a visual frame at the current step. Furthermore, guided by our training reward (below in Sec.~\ref{sec:training}), \SM{the previously observed visual frames} and audio together enable our policy to \SM{anticipate the current frames and skip them if they are deemed redundant,}
thereby improving mapping accuracy for a low visual budget. 

\vspace*{-0.15in}
\paragraph{Policy network.}
The policy network consists of a GRU that estimates an updated history $h_t$ 
along with the current state representation $g_t$, 
using the fused embedding $e_{t}$ and the history of states $h_{t-1}$.
An actor-critic module takes $g_t$ and $h_{t-1}$ as inputs and predicts a policy distribution $\pi_\theta(a_{i, t} | g_t, h_{t-1})$ per ego along with the value of the state $H_{\theta}(g_t, h_{t-1})$ ($\theta$ are policy parameters). 
The
\SM{policy} samples an action $a_{i, t} \in \big\{0, 1\big\}$ for every $E_i$.
$a_{i,t}=1$ corresponds to selecting $\mathcal{V}_{i, t}$, \KGcr{whereas $a_{i,t}=0$ means to skip that frame}.

\subsection{Shared scene mapper}\label{sec:mapper}
\SM{Whereas $O^\pi(t)$ denotes our policy input (Sec.~\ref{sec:policy}), $O^M(t)$ denotes the input to our shared scene mapper $f^M$ at step $t$, such that}
$O^M(t) = \big\{(V_{i, j}, S_{i, j}, S^{'}_{i, j}, P_{i, j}, P^{'}_{i, j}): i = 1 \ldots 2, j = 1 \ldots t\big\}$.
$f^m$ starts by embedding each component of $O^M(t)$ using a separate network. 
This is followed by a multi-modal memory that stores the embeddings since the start of the episode. 
Finally, a transformer~\citep{vaswani2017attention} predicts an estimate $\tilde{M}(t)$ of the scene map conditioned on the multi-modal memory 
and the egos' poses in the episode.

\vspace*{-0.15in}
\paragraph{Multi-modal mapper embedding.}
For the visual input $V_{i, j}$, we encode $V^R_{i, j}$ and $V^M_{i, j}$ using separate CNNs 
and do a channel-wise concatenation to get visual features $\hat{v}_{i,j}$.
Similarly speech is encoded using separate CNNs to get $\hat{s}_{i, j}$ and $\hat{s}^{'}_{i, j}$. 
\SM{Each of $\hat{v}$, $\hat{s}$ and $\hat{s}^{'}$}
is of size $4 \times 4 \times 1024$. 

For both vision and speech, we compute two positional embeddings, $p^{\RN{1}}$ and $p^{\RN{2}}$. 
They encode the pose of the egos in the 3D space, 
and the index of each 1024-dimensional feature in the visual or speech features in the raster order respectively. 
Whereas $p^{\RN{1}}$ helps discover spatial cues as a function of the egos' location in the 3D scene, 
$p^{\RN{2}}$ enables our model to attend to different modalities in a more fine-grained manner.
For both, we compute an 8-dimensional sinusoidal positional encoding~\citep{vaswani2017attention} and then pass it through a linear layer to obtain a 1024-dimensional embedding. 
For $p^{\RN{2}}$, we additionally repeat this process for every feature index in the raster order. 
Lastly, we reshape $p^\RN{1}$ and add it with $p^\RN{2}$ to produce $4 \times 4 \times 1024$-dimensional positional embeddings,  $\hat{p}_{i, j}$ for $\hat{v}_{i, j}$ and $\hat{s}_{i, j}$, and $\hat{p}^{'}_{i, j}$ for $\hat{s}^{'}_{i, j}$.

Following~\citep{Majumder2022FewShotAL}, we also learn an embedding $\hat{m}_{i, j} \in \big\{\hat{m}_V, \hat{m}_S, \hat{m}_{S'}\big\}$ 
to capture different modality types, where $\hat{m}_V$ represents vision, and $\hat{m}_S$ and $\hat{m}_{S'}$ represent the speech from self and that of the other ego, respectively. 
The modality-based embeddings help our model differentiate between different modalities 
and better map the scene by learning complementary spatial cues from them. 

\vspace*{-0.15in}
\paragraph{Multi-modal memory.}
For the visual input $V_{i, j}$, we add its embedding $\hat{v}_{i, j}$ with its positional embedding $\hat{p}_{i , j}$ and modality embedding $\hat{m}^V_{i, j}$, and flatten the sum to get a $16 \times 1024$-dimensional embedding. Similarly, we fuse the speech embeddings by taking their sum and flattening them. This generates a multi-modal memory of fused embeddings $o$, such that $o = \big\{o^V_{1, 1}, \ldots, o^V_{2, t}, o^S_{1, 1}, \ldots, o^S_{2, t}, o^{S'}_{1, 1}, \ldots, o^{S'}_{2, t}\big\}$.

\vspace*{-0.15in}
\paragraph{Occupancy prediction.}
To predict the underlying scene occupancy, we first use a transformer encoder~\citep{vaswani2017attention} to attend to the embeddings in $o$ and capture short- and long-range correlations within and across modalities using a stack of self-attention layers. This generates an audio-visual representation that models the spatial layout of the 3D scene. 

Next, we use a transformer decoder~\citep{vaswani2017attention} to perform cross-attention on the audio-visual representation of the scene conditioned on the embedding 
$\hat{p}_{i, j}$ for every pose $P_{i, j}$ in $O^M(t)$ and generate an embedding $d_{i, j}$ for the pose. Finally, we upsample $d_{i, j}$ using a multi-layer network $U$ comprising transpose convolutions and a sigmoid layer at the end to predict an estimate $\tilde{M}_{i, j}$ of the ground-truth local $360^\circ$ FoV map for the pose, $M_{i, j}$. Both $M_{i, j}$ and its estimate $\tilde{M}_{i, j}$ are two-channel binary occupancy maps of size $H \times W$.
To obtain the estimated map $\tilde{M}(t)$ for the scene,
we register each prediction $\tilde{M}_{i, j}$ onto a larger shared map using the pose $P_{i, j}$ and threshold the final shared map at $0.5$ (see Supp. for map registration details).
Importantly, the shared map allows communication between both egos' data streams for more informed mapping and sampling, as we show in results.

\subsection{Model training}\label{sec:training}

\paragraph{Policy training.} 
We propose a novel dense RL reward 
to train policy $\pi^V$:
\begin{equation*}
    r(t) = \Delta Q(t) - \eta \text{ } * \text{ } \rho(t).
\end{equation*}
$\Delta Q(t)$ measures the improvement \SM{in mapping} from taking actions $\big\{a_{i,t}: i = 1 \ldots 2\big\}$ 
over not sampling any visual frame at step $t$.
$\rho(t)$ is a penalty term to discourage sampling \SM{a frame from the same pose}
more than once, which we weight by $\eta$.
We define $\Delta Q(t)$ as
\begin{equation*}
\Delta Q(t) = Q\big(\tilde{M}(t) \text{ } | \text{ } O^M(t)\big) - Q\big(\tilde{M}(t) \text{ } |  \text{ } (O^M(t) \text{ } \backslash \text{ } V_t)\big),
\end{equation*} where $Q$ is a map quality measure, $Q(X | Y)$ represents the quality of map estimate $X$ given inputs $Y$, and $(O^M(t) \text{ } \backslash \text{ } V_t)$ denotes the mapper inputs devoid of any visual frame for the current step. 
We define $\rho(t)$ as
\begin{equation*}
    \rho(t) = \sum_{i = 1\ldots 2} \text{ } a_{i, t} * \mathbbm{1}(V_{i, t} \in O^M(t - 1)), 
\end{equation*} where the indicator function checks if $V_{i, t}$ was used in mapping before. 
While $\Delta Q(t)$ incentivizes sampling frames that provide a big boost to the mapping accuracy over skipping them, 
$\rho(t)$ penalizes wasting the visual budget on redundant sampling, 
thereby maximizing mapping performance within the constraints of a limited budget.
We set $\rho=0.03$ in all our experiments and define $Q$ as the average F1 score over the occupied and free classes in a predicted occupancy map. 
\SMC{We empirically observe that alternate penalty terms based on visual similarity, like the ORB~\citep{6126544} feature matching similarity, do not improve the model performance.}

We train $\pi^V$ with Decentralized Distributed PPO (DD-PPO)~\citep{Wijmans2020DDPPOLN}. 
The DD-PPO loss consists of a value loss, policy loss and an entropy loss to promote exploration (see Supp. for details).

\vspace*{-0.15in}
\paragraph{Mapper training.} At each step $t$, we train the shared mapper $f^m$ with a 
loss $\mathcal{L}^M(t)$, such that 
\begin{equation*}
   \mathcal{L}^M(t) = \frac{1}{2 \times t}\sum_{i = 1 \ldots 2}\sum_{j = 1 \ldots t} \text{BCE}(\tilde{M}_{i, j}, M_{i, j}),
\end{equation*} where \SM{$\text{BCE}(\tilde{M}_{i, j}, M_{i, j})$ is the average binary cross entropy loss between $\tilde{M}_{i, j}$ and $M_{i, j}$.}

\vspace*{-0.15in}
\paragraph{Training curriculum.}
To train
our model, we first pretrain mapper $f^m$ in two phases 
and then train the policy $\pi^V$ while keeping $f^m$ frozen. In 
phase 1, we train $f^m$ without visual sampling, \ie all visual frames 
are provided at each step. In 
phase 2, we finetune the pretrained weights of $f^m$ from 
phase 1 on episodes where we randomly drop views to satisfy the budget $B$. While phase 1
improves convergence when training with visual sampling, 
phase 2 helps with reward stationarity when training our RL policy.   
\KGcr{Both phases use the same training, validation and testing data.}

%% file: sections/experiments.tex
\section{Experiments}
\label{sec:experiments}

\paragraph{Experimental setup.}
For our main experiments, we use
SoundSpaces~\citep{chen2020soundspaces} acoustic simulations with AI-Habitat~\citep{savva2019habitat} and Matterport3D~\citep{chang2017matterport3d} visual scenes. 
While Matterport3D provides dense 3D meshes and image scans of real-world houses and other indoor scenes, 
SoundSpaces provides room impulse responses (RIRs) at a spatial resolution of 1m for Matterport3D that model 
all real-world acoustic phenomena~\citep{chen2020soundspaces}. 
This setup allows us to evaluate with as many as 83 scenes, split in 56/10/17 for train/val/test, 
compare against relevant prior work~\citep{ramakrishnan2020occupancy, purushwalkam2021audio} and report reproducible results. 

We also collect real-world data in 
a mock-up apartment due to the absence of a publicly available alternative suited for our task. 
We capture a dense set of RGB images using a Samsung S22 camera and generate the corresponding depth images 
using monocular depth estimation~\citep{eftekhar2021omnidata, kar20223d}. 
To compute the RIRs, following~\citep{Farina2000SimultaneousMO}, we generate a sinusoidal sweep sound from 20Hz-20kHz 
with a loudspeaker at source location, capture it with an Eigenmike at a receiver location, 
and convolve the spatial sound with the inverse of the sweep sound to retrieve the RIR. 
All capturing devices are placed at a height of 1.5 m. 
We generate occupancy maps by backprojecting the depth images (cf. Sec.~\ref{sec:inp_prep}) 
and register them onto a shared top-down map before taking egocentric crops to generate the local occupancy inputs and targets. 

Note that \emph{both datasets have real-world visuals} as they are captured in the real environments; SoundSpaces has \KGcr{high-quality} simulated audio while the apartment data has real-world collected audio RIRs.

\vspace*{-0.1in}
\paragraph{Conversation episode.}
For each episode,(simulated or real),
we randomly place the egos in a scene. 
Episode length is $T = 16$ and $8$ for simulation and real, respectively. 
At each step, the egos execute a movement from $\mathcal{A} = \big\{MoveForward, TurnLeft, TurnRight\big\}$, where $MoveForward$ moves an ego forward by $1$ m, and 
the $Turn$ actions rotate the ego by $90^\circ$\SMC{, as required by SoundSpaces}. Further, either of the egos speaks or both speak with equal probability of $\frac{1}{3}$ at every step, 
i.e., there are no moments of \KGcr{total} silence, \KGcr{but one ego may be silent}. 
The egos stay between $1-3$m from each other \SMC{(typical in a conversation)}
so that they do not collide 
and 
each ego is audible by the other at all times. 
This results in train/val
splits of 1,955,334/100 episodes in simulation, and a simulated/real-world test split of 1000/27 episodes. \SMC{See Supp. for additional analysis on the model performance vs. training data size.}
We set the visual budget to $B=2$ for our main experiments (see Supp.~for $B=4,6$ evaluations).  
Note that these episodes are \KGcr{solely} to generate video data; our task requires processing passive video, not controlling embodied agents.

\vspace*{-0.15in}
\paragraph{Observations and model output.}
For the occupancy maps,  
\KGcr{consistent with~\citep{chen2019learning, Chaplot2020Learning, ramakrishnan2020occupancy}},
we generate $(31 \times 31 \times 2)$-dimensional input maps that cover $3.1 \times 3.1$ m$^2$ in area at a resolution of 0.1 m, and set the local target map size to 
$H \times W = 6.4 \times 6.4$ m$^2$ ($\sim41$ m$^2$).
For speech, we use 100 distinct speakers from LibriSpeech~\citep{7178964}, split in 80/11 for \emph{heard}/\emph{unheard}, where \emph{unheard} speech is only used in testing. We assume camera poses, as modern AR devices are equipped with motion sensors that can robustly estimate relative poses~\citep{liu2020tlio}. 
We test our robustness to ambient sounds that get mixed with the egos' speech, and incorporate odometry noise models~\cite{ramakrishnan2020occupancy, purushwalkam2021audio} in Supp.

\vspace*{-0.15in}
\paragraph{Evaluation settings.}
We evaluate our model in two settings: 1) \emph{passive mapping},
the mapper has access to all visual frames in an episode (i.e., the camera is always on), and 2) \emph{active mapping}, where the mapping agent has to actively sample frames to meet the visual budget $B$. This 
helps disentangle our modeling contributions---whereas \emph{passive mapping} lets us show improvements in the mapper $f^M$ 
over existing
methods~\citep{ramakrishnan2020occupancy, purushwalkam2021audio}, \emph{active mapping} helps demonstrate the benefits of smart visual sampling.

We use standard evaluation metrics~\citep{ramakrishnan2020occupancy}: \textbf{F1 score} and \textbf{IoU} (intersection over union) between the predicted and target scene maps. For both metrics, we report the mean over the free and occupied classes.
For \emph{active mapping}, we average the metrics
over 3 random 
seeds.
We use the following baselines to compare our model`s efficacy.

\noindent\emph{Passive mapping:} 
\begin{itemize}[leftmargin=*,topsep=0pt,partopsep=0pt,itemsep=0pt,parsep=0pt]
    \item \textbf{All-occupied:} a naive baseline that predicts all locations in its map estimate as occupied
    \item \textbf{Register-inputs:} a naive baseline that registers the input maps onto a shared map and uses it as its prediction
    \item \textbf{OccAnt~\citep{ramakrishnan2020occupancy}:} a vision-only 
    SOTA model that uses the RGB-D images at each step to anticipate the occupancy of the area around an ego that's outside its visible range.
    \item \textbf{AV-Floorplan~\citep{purushwalkam2021audio}:} an audio-visual SOTA
    model that \emph{passively} predicts the floorplan of a scene using a walkthrough in it, where the audio is either self-generated or comes from semantic sources in the scene. We adapt the model for our occupancy prediction task \SM{and give it the exact same audio-visual observations as our model}.  
\end{itemize}
\noindent\emph{Active mapping:}
\begin{itemize}[leftmargin=*,topsep=0pt,partopsep=0pt,itemsep=0pt,parsep=0pt]
    \item \textbf{Random:} an agent that selects visual frames randomly for each ego as long as the budget allows
    \item \textbf{Greedy:} an agent that greedily uses up the visual budget by sampling frames as early as possible 
    \item \textbf{Unique-pose:} an agent that samples a frame for every new \SM{ego} pose in the episode 
\end{itemize}
In \emph{active mapping}, we use the model from the second pretraining phase (Sec.~\ref{sec:training})
as the mapper for all models for fair comparison. Thus, any difference in performance is due to the quality of each method's sampling decisions. 

\indent See Supp. for all other details like network architectures, training hyperparameters, and baseline implementation.

\subsection{Map prediction results}\label{sec:map_pred_results} 

\begin{table}[t]
  \centering
    \scalebox{0.82}{
    \setlength{\tabcolsep}{2pt}
    \begin{tabular}{lcc|cc}
    \toprule
                    &   \multicolumn{2}{c|}{Simulation} &  \multicolumn{2}{c}{Real world}\\
    Model           & {F1 score $\uparrow$} & {IoU $\uparrow$} & {F1 score $\uparrow$} & {IoU $\uparrow$} \\    \midrule
    All-occupied     &63.4 & 48.8   &  36.2 &  23.8\\
    Register-inputs     & 72.6  & 60.1 & 50.8 &  35.0\\   
    OccAnt~\citep{ramakrishnan2020occupancy}   & 74.5 & 62.7 & 53.9 & 38.3 \\   
    AV-Floorplan~\citep{purushwalkam2021audio}   & 79.3 & 67.9 & 54.5 & 38.7  \\   
    \textbf{Ours}                       & \textbf{81.8} & \textbf{71.4} & \textbf{55.5} & \textbf{39.2} \\
    \midrule
    Ours w/o vision      & 72.8 & 60.3 & 50.8 & 35.0 \\ 
    Ours w/o audio      & 78.1 & 66.7 & 54.1 & 38.0 \\    
    Ours w/o $E^{'}_{i}$'s speech      &  81.5 & 70.9  & 55.4 & 39.1\\    
    Ours w/o shared mapping      & 80.7 & 70.0  & 54.9 & 38.6 \\
    Ours w/o modality tag & 79.1 & 67.8 & 54.5 & 38.1\\
    \bottomrule
  \end{tabular}
}
  \vspace*{-0.25cm}
  \caption{\emph{Passive mapping} performance ($\%$).  All our gains are statistically significant ($p$ $\leq 0.05$).} 
  \label{table:pm_main}
\vspace{-0.3cm}
\end{table}

\paragraph{Passive mapping.}
Table~\ref{table:pm_main} (top) reports the prediction quality of all models in the \emph{passive mapping} setting. Naive baselines (All-occupied, Register-inputs) perform worse than the learned models, showing the complexity of our map prediction task. AV-Floorplan~\citep{purushwalkam2021audio} fares the best among all baselines. Its improvement over OccAnt~\citep{ramakrishnan2020occupancy} demonstrates the benefits of exploiting the spatial cues in audio for mapping and using an attention-based model to leverage the long- and short-range correlations in the audio-visual inputs. 

Our method outperforms all baselines. 
Its improvement over AV-Floorplan~\citep{purushwalkam2021audio} underlines the efficacy of performing attention at different granularities---across modalities, within a single modality and within a single input---guided by our positional and modality type embeddings. 
It also generalizes to the real-world setting and retains its benefits over the baselines, 
even without retraining on the real-world data. 
However, we do observe a drop in performance gains, probably due to the large sim-to-real gap.

\begin{figure}[t]
    \centering
    \begin{subfigure}[b]{0.49\linewidth}
    \centering
    \includegraphics[width=\linewidth]{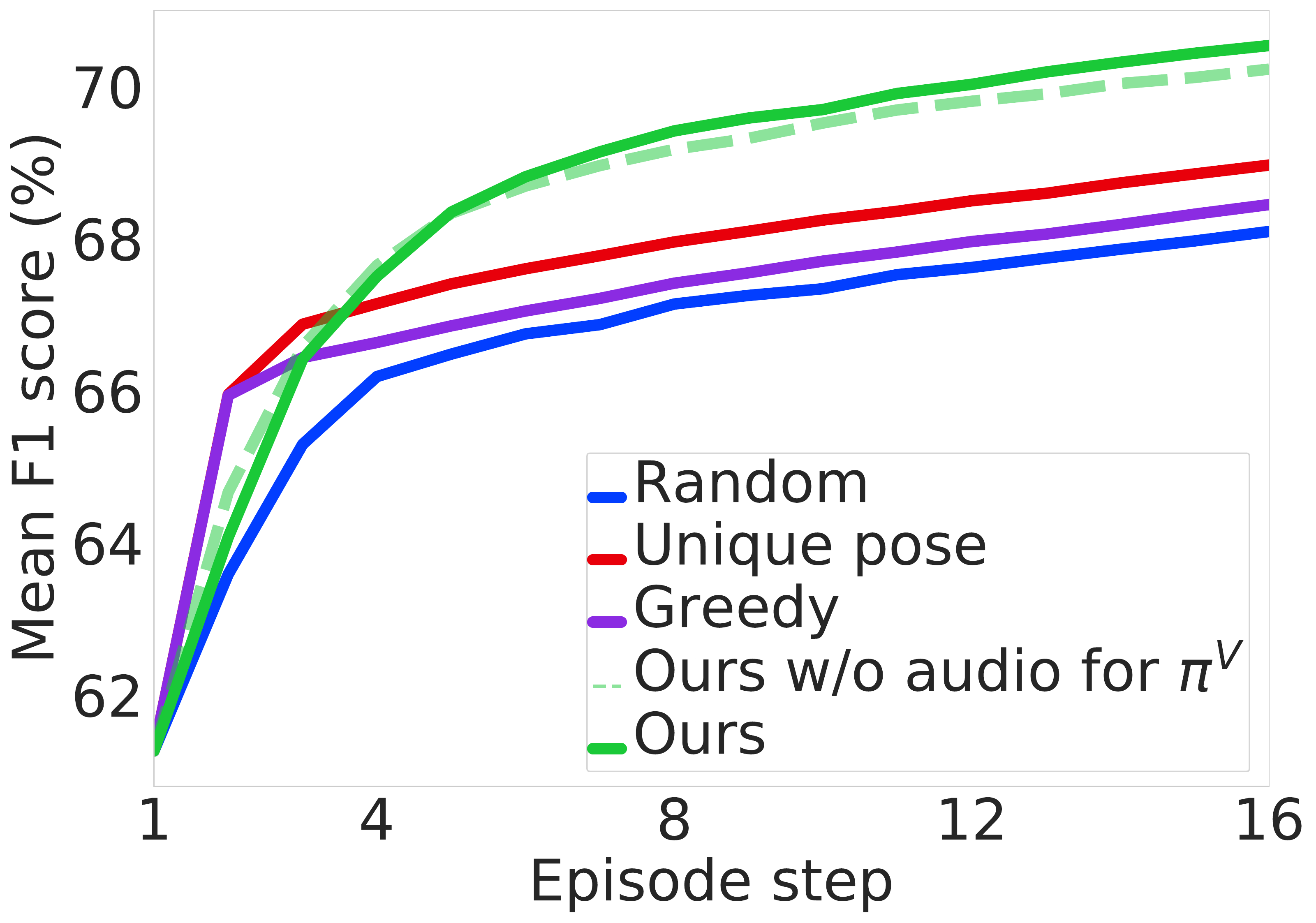}
    \caption{Simulation}
    \label{fig:am_tgt64_b2}
    \end{subfigure}\hfill
    \begin{subfigure}[b]{0.49\linewidth}
    \centering
    \includegraphics[width=\linewidth]{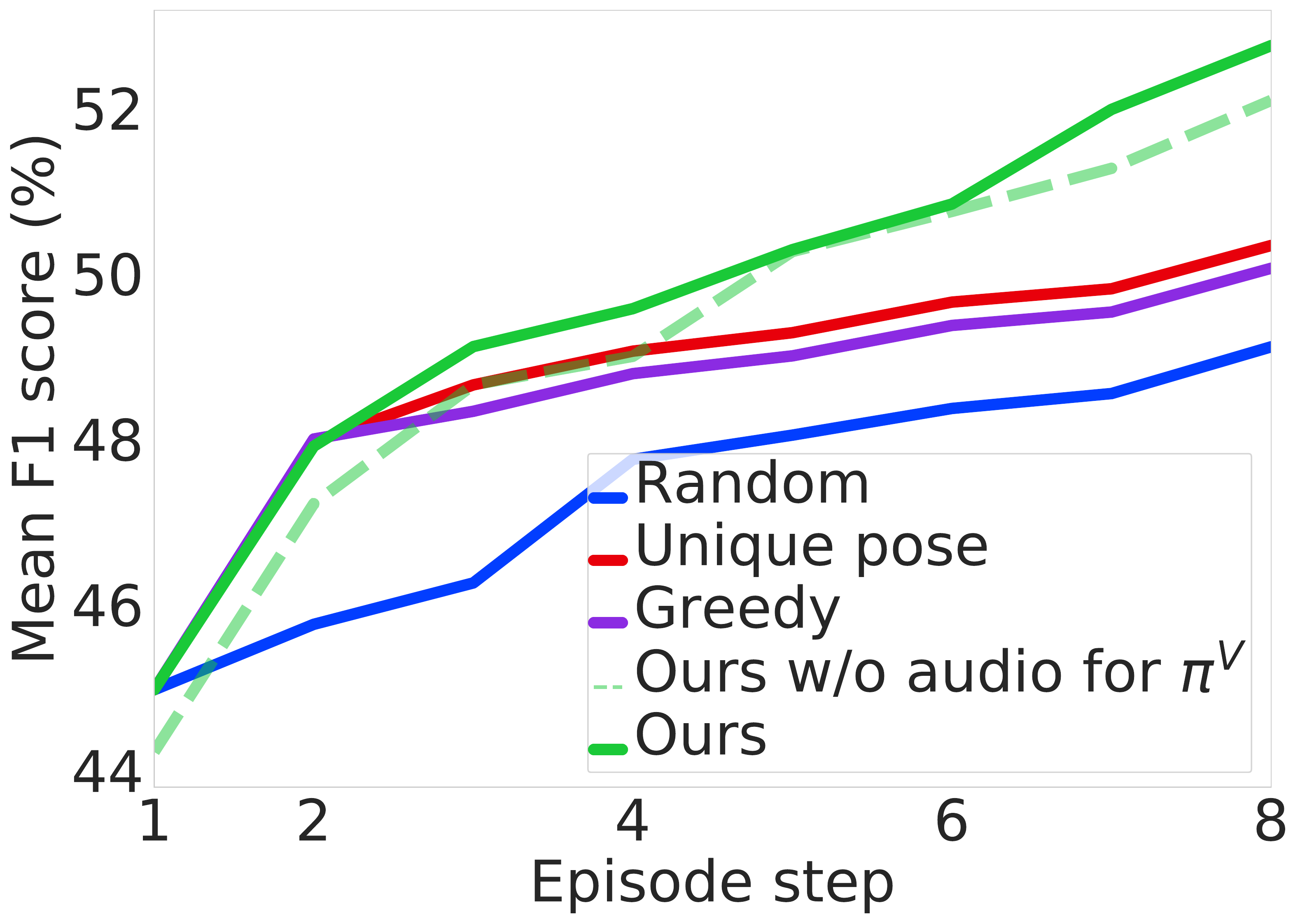}
    \caption{Real world}
    \label{fig:am_rw_tgt64_b2}
    \end{subfigure}\hfill
\caption{
\emph{Active mapping} performance vs. episode step.
}
\label{fig:am_b2}
\end{figure}

\begin{figure}[t]
    \centering
    \includegraphics[width=0.6\linewidth]{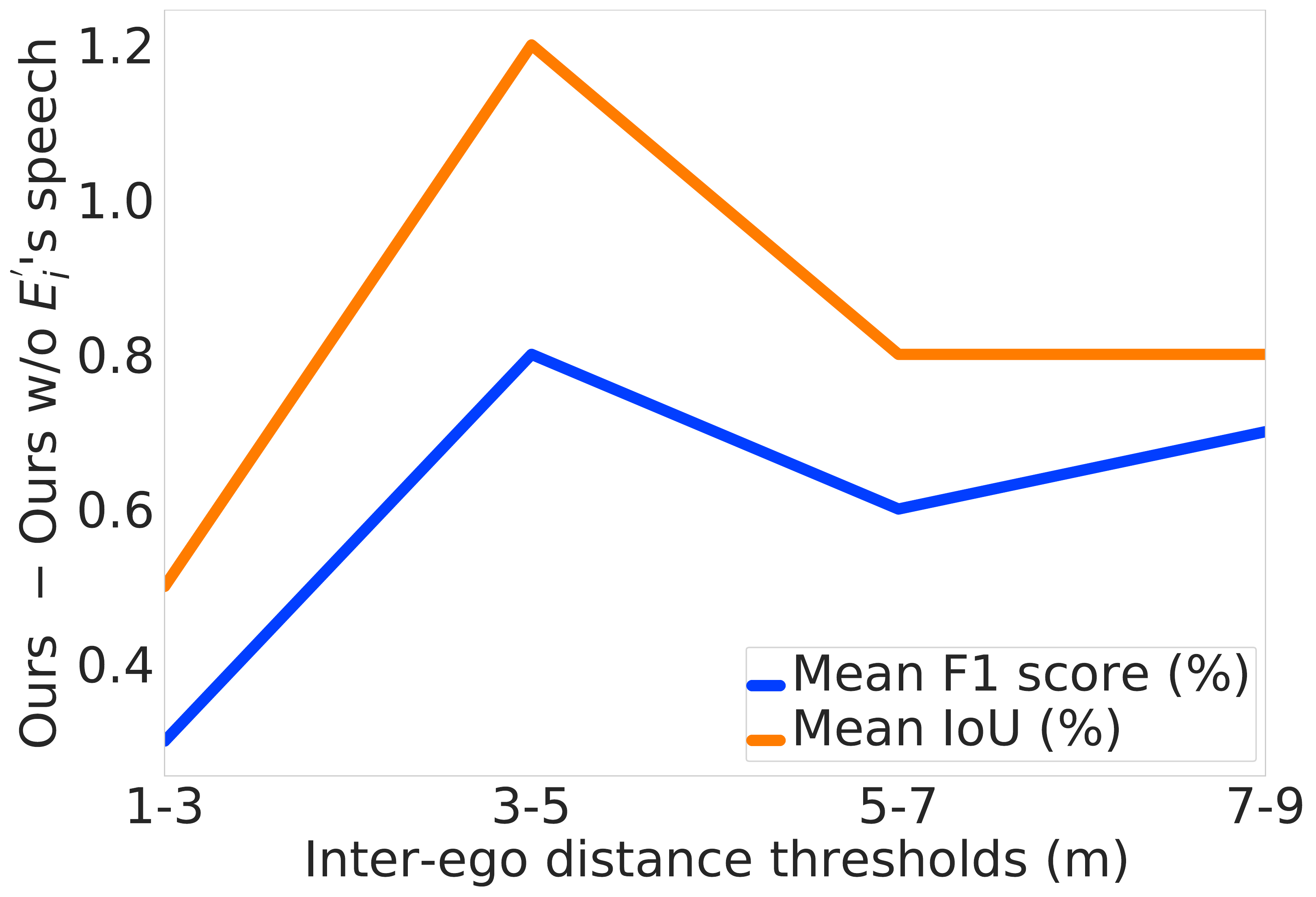}
\caption{\SM{Impact of the other ego's speech on \emph{passive mapping}
vs. distance between the egos.}
} 
    \label{fig:am_tgt64_b4_ours_vs_oursWoFrFld_dffIntrEgDsts}
    \vspace{-0.3cm}
\end{figure}

\begin{figure*}[!t] 
    \centering
    \includegraphics[width=\linewidth]{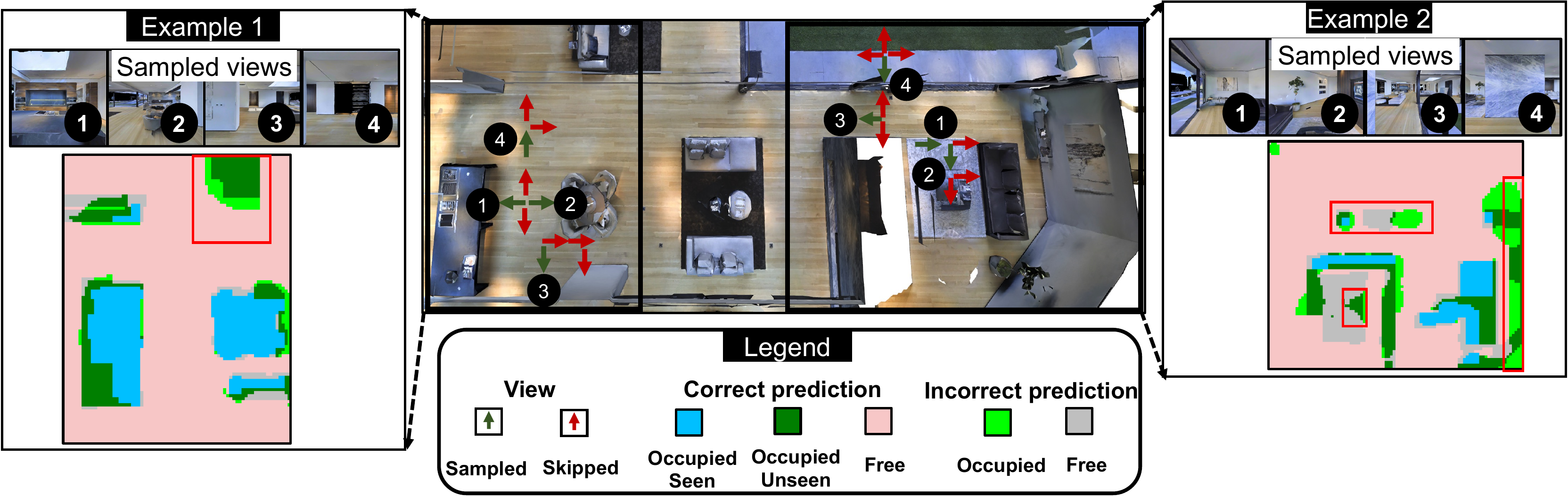}
    \caption{
    \SM{Sample episodes for our \emph{active mapping} model. While our policy samples only the salient visual frames, our mapper can both complete partially seen objects as well as anticipate objects never seen before in the sampled visuals (red boxes on the maps).}
    }
\label{fig:qual_results}
\vspace{-.25cm}
\end{figure*}

\vspace*{-0.15in}
\paragraph{Active mapping.}
Fig.~\ref{fig:am_b2} shows the active mapping performance as a function of episode progress. 
Employing naive heuristics for sampling, like Random or Greedy, is
\KGcr{insufficient} for high-quality mapping, emphasizing the high levels of redundancy in the visual frames. Unique-pose improves over both Random and Greedy, 
showing that sampling diverse viewpoints provides more information about the underlying scene geometry. 

Although the baselines make initial progress, they flatten quickly and our model eventually outperforms them all, on both real-world and simulated data (\KGcr{gains are again statistically significant, $p \leq 0.05$}). 
This highlights the benefits of learning a smart policy that,
given the audio streams and its past visual samples, understands the value of sampling a visual frame for mapping by taking cues from our novel reward.
Besides, on the real-world data,we see better performance margins over the baselines towards the end of episodes,
showing that our policy can adaptively defer visual sampling 
to improve mapping. 
\SM{Owing to smart sampling, the per-episode \emph{reduction} in \SMC{the capture / }processing \SMC{cost} for $B=2$ is \SMC{74 Watts (W) / }7.2 GFLOPS in simulation and \SMC{37 W / }3.6 GFLOPS in the real world \KGcr{video}.} \SMC{See Supp. for further analysis of our model's power and time cost.}

\subsection{Model analysis}\label{sec:mod_analysis}
\paragraph{Ablations.}
In Table~\ref{table:pm_main} (bottom), we ablate the components of our model for \emph{passive mapping}. Upon removing audio, our model experiences a large drop in mapping performance, which indicates that our model leverages complementary spatial cues in audio and vision. We also see a drop in the map quality  when our model \KGcr{lacks} access to the speech from the other ego ($E^{'}_i$). This shows that $E^{'}_i$'s speech can better reveal the more global scene geometry than $E_i$'s own speech.n Fig.~\ref{fig:am_tgt64_b4_ours_vs_oursWoFrFld_dffIntrEgDsts} further shows that the impact of the other ego's speech becomes more prominent for larger inter-ego distances ($3-5$ m vs. $1-3$ m), in which case the two types of speech are dissimilar enough to carry complementary geometric cues, but reduces for even larger distances (5 m or more), in which case $E^{'}_i$ is too far for its speech to carry useful cues about $E_i$'s local scene geometry. Moreover, unlike the ablation that does not perform shared mapping, our model benefits significantly from \emph{jointly attending} to the observations of the egos and exploiting the complementary information in them
---even though both models use the exact same audio-visual observations, including both speech from self and the other ego. \SMC{\KGcr{Finally,} without our modality type embeddings, our model does worse, showing their role in learning complementary features from vision and audio.}

For \emph{active mapping}, Fig.~\ref{fig:am_b2} shows a drop in the 
mapping performance upon removing audio from the policy inputs. This implies that our policy exploits audio to reason about the level of redundancy in a new visual frame and improve the mapping quality vs. visual budget tradeoff. 
On the more challenging real-world setting, audio plays an even bigger role, as shown by the larger performance drop in Fig.~\ref{fig:am_rw_tgt64_b2}. 

See Supp. for 
\textbf{similar results} with
\SMC{1) ambient sounds,} 2) \emph{unheard} speech, 3) higher budgets, 4) sensor noise, 5) larger target maps\SMC{, 6) different ego initializations, and 7) multiple random data splits}.

\vspace*{-0.15in}
\paragraph{Qualitative results.}
Fig.~\ref{fig:qual_results} shows two successful \emph{active mapping} episodes of our method. 
Note how our model samples views that tend have to little visual overlap 
but are informative of the surrounding geometry (both occupied and free spaces). 
Besides, it is able to complete structures only partially visible in the sampled views,
and more interestingly, leverage the synergy of audio and vision to anticipate unseen areas (red boxes on the occupancy maps in Fig.~\ref{fig:qual_results}).

\vspace*{-0.15in}
\paragraph{Failure cases.}
We notice two common 
failure cases 
with \emph{active mapping}:
episodes where the egos stay at the same location, leading to very few informative visual frames 
to sample from; and 
episodes with highly unique visual samples at every trajectory step, 
in which case 
each sample is useful and our model behaves similar to Unique-pose or Greedy.
For \emph{passive mapping}, our model fails 
with very complex scenes that commonly have objects in spaces where both vision and audio cannot reach (\eg narrow corners)

%% file: sections/conclusion.tex
\vspace*{-0.05in}
\section{Conclusion}
\vspace*{-0.05in}

We introduce Chat2Map, a new task aimed at scene mapping using audio-visual feeds from egocentric conversations.
We develop a novel approach for Chat2Map  comprising a shared scene mapper 
and a visual sampling policy based on a novel reinforcement learner that smartly samples the visuals only when necessary.
We show promising performance on both simulated and real-world data from over 80 scenes. 

\small{\SMC{\noindent \textbf{Acknowledgements:} UT Austin is supported in part by the IFML NSF AI  Institute.  KG is paid as a research scientist by Meta.}}

%% file: sections/supp.tex
\section{Supplementary Material}
In this supplementary material we provide additional details about:
\begin{itemize}
    \item Video (with audio) for qualitative illustration of our task and qualitative assessment of our map predictions (Sec.~\ref{sec:supp_video})

    \item \SMC{Societal impact of our work (Sec.~\ref{sec:supp_societal_impact})}

    \item \SMC{Detailed analysis of the power and time cost of our model (Sec.~\ref{sec:supp_cost}), as mentioned in Sec.~\ref{sec:map_pred_results}}

    \item \SMC{Experiment to show the effect of ambient environment sounds on mapping accuracy (Sec.~\ref{sec:supp_ambient_sounds}), as referenced in Sec.~\ref{sec:mod_analysis}}
    
    \item Experiment to show the effect of \emph{unheard} sounds on map predictions (Sec.~\ref{sec:supp_unheard}), as noted in Sec.~\ref{sec:mod_analysis} 
    
    \item Experiment to show the impact of the visual budget $B$ (Sec.~\ref{sec:task}) on mapping quality (Sec.~\ref{sec:supp_B_value}), as referenced in Sec.~\ref{sec:experiments} and~\ref{sec:mod_analysis}. 
    
    \item Experiment to show the effect of sensor noise on mapping accuracy (Sec.~\ref{sec:supp_sensor_noise}), as mentioned in Sec.~\ref{sec:experiments} and~\ref{sec:mod_analysis}. 
    
    \item Experiment to show mapping performance as a function of the target map size (Sec.~\ref{sec:supp_target_map_size}), as noted in Sec.~\ref{sec:mod_analysis}. 

    \item \SMC{Experiment to show the effect of different types of ego initializations on map predictions (Sec.~\ref{sec:supp_ego_inits}), as referenced in Sec.~\ref{sec:mod_analysis}}

    \item \SMC{Experiment to show the effect of multiple random dataset splits on mapping quality (Sec.~\ref{sec:supp_mult_data_splits}), as referenced in Sec.~\ref{sec:mod_analysis}}

    \item \SMC{Experiment to show the effect of the training data size on the model performance (Sec.~\ref{sec:supp_train_data_size}), as mentioned in Sec.~\ref{sec:experiments}}
    
    \item Dataset details (Sec.~\ref{sec:supp_dataset}) in addition to what's provided in Sec.~\ref{sec:experiments}.
    
    \item Additional baseline details for reproducibility (Sec.~\ref{sec:supp_baselines}), as referenced in Sec.~\ref{sec:experiments}.
    
    \item Architecture and training details (Sec.~\ref{sec:supp_arc_n_train}), as noted in Sec.~\ref{sec:experiments}.
    
\end{itemize}

\subsection{Supplementary video}\label{sec:supp_video}
The supplementary video qualitatively depicts our task, Chat2Map:Efficient Scene Mapping from Multi-Ego Conversations. Moreover, we qualitatively show our model’s mapping quality by comparing the predictions against the ground truths and the visual samples chosen by our sampling policy for efficient mapping\SMC{, and analyze common failure modes of our model}. We also demonstrate the acoustically realistic SoundSpaces~\citep{chen2020soundspaces} audio simulation platform that we use for our core experiments. Please use headphones to hear the spatial audio correctly. \SMC{The video is available on \url{http://vision.cs.utexas.edu/projects/chat2map}.}

\SMC{
\subsection{Societal impact}\label{sec:supp_societal_impact}
Our model enables efficiently mapping a scene from natural conversations. This has multiple applications with a positive impact. For example, accurate mapping of an unseen environment enables many downstream applications in AR/VR (e.g., accurate modeling of scene acoustics for an immersive user experience) and robotics (e.g., a robot using a scene map to better navigate and interact with its environment). However, our model relies on speech inputs, which when stored or used without sufficient caution could be prone to misuse by unscrupulous actors. Besides, the dataset used in our experiments contains indoor spaces that are predominantly of the Western design, and with a certain object distribution that is common to such spaces.
This may bias models trained on such data toward similar types of scenes and reduce generalization to scenes from other cultures. More innovations in the model design to handle strong shifts in scene layout and object distribtutions, as well as more diverse datasets are needed to mitigate the impact of such possible biases.

\subsection{Power and time cost}\label{sec:supp_cost}
With visual budget $B=2$ and episode length $T=16$ (Sec.~\ref{sec:task} and~\ref{sec:experiments}), our model skips 28 frames and saves 7.2 GFLOPs in mapping but adds 24.1 GFLOPs due to the policy, thus adding a net of 16.1 GFLOPs. 
This translates to around 0.5 Watt~\citep{desislavov2021compute}
of extra power for running the active mapper but a saving~\citep{likamwa2014draining}
of $\sim 28 \times 3 = 74$ Watt in camera capture, thus saving a net of $\sim$73.5 Watt in power. The total runtime of our model in this setting is 5.6 s on a Quadro GV100 GPU.

Compared to the heuristical baseline policies (Sec.~\ref{sec:experiments}), our policy adds $\sim$24GFLOPs, consuming 1 extra Watt~\citep{desislavov2021compute} on modern GPUs, while improving the map by 14 m$^2$ on avg (Fig.~\ref{sec:task}).

\subsection{Ambient and background sounds}\label{sec:supp_ambient_sounds}
\begin{table}[t] 
  \centering
    \scalebox{0.85}{
    \setlength{\tabcolsep}{4pt}
    \begin{tabular}{lcc}
    \toprule
    Model           & {F1 score $\uparrow$} & {IoU $\uparrow$} \\    \midrule
    All-occupied     & 63.4 & 48.8  \\
    Register-inputs     & 72.6 & 60.1\\
    OccAnt~\citep{ramakrishnan2020occupancy}   & 74.5 & 62.7 \\
    AV-Floorplan~\citep{purushwalkam2021audio}   &  78.7 & 67.5 \\
    \textbf{Ours}                       & \textbf{81.9} & \textbf{71.5}\\
    \midrule
    Ours w/o vision      & 73.5  & 61.2 \\
    Ours w/o audio      & 78.1  & 66.7 \\
    Ours w/o $E^{’}_{i}$‘s speech      & 81.5 & 70.9 \\
    Ours w/o shared mapping      & 80.0  & 69.1 \\
    \bottomrule
  \end{tabular}
  }
  \caption{\emph{Passive mapping} performance ($\%$)
  with ambient
  sounds.
  }
  \label{table:pm_amb}
\vspace{-0.25cm}
\end{table}

\begin{figure}[t]
    \centering
    \includegraphics[width=0.8\linewidth]{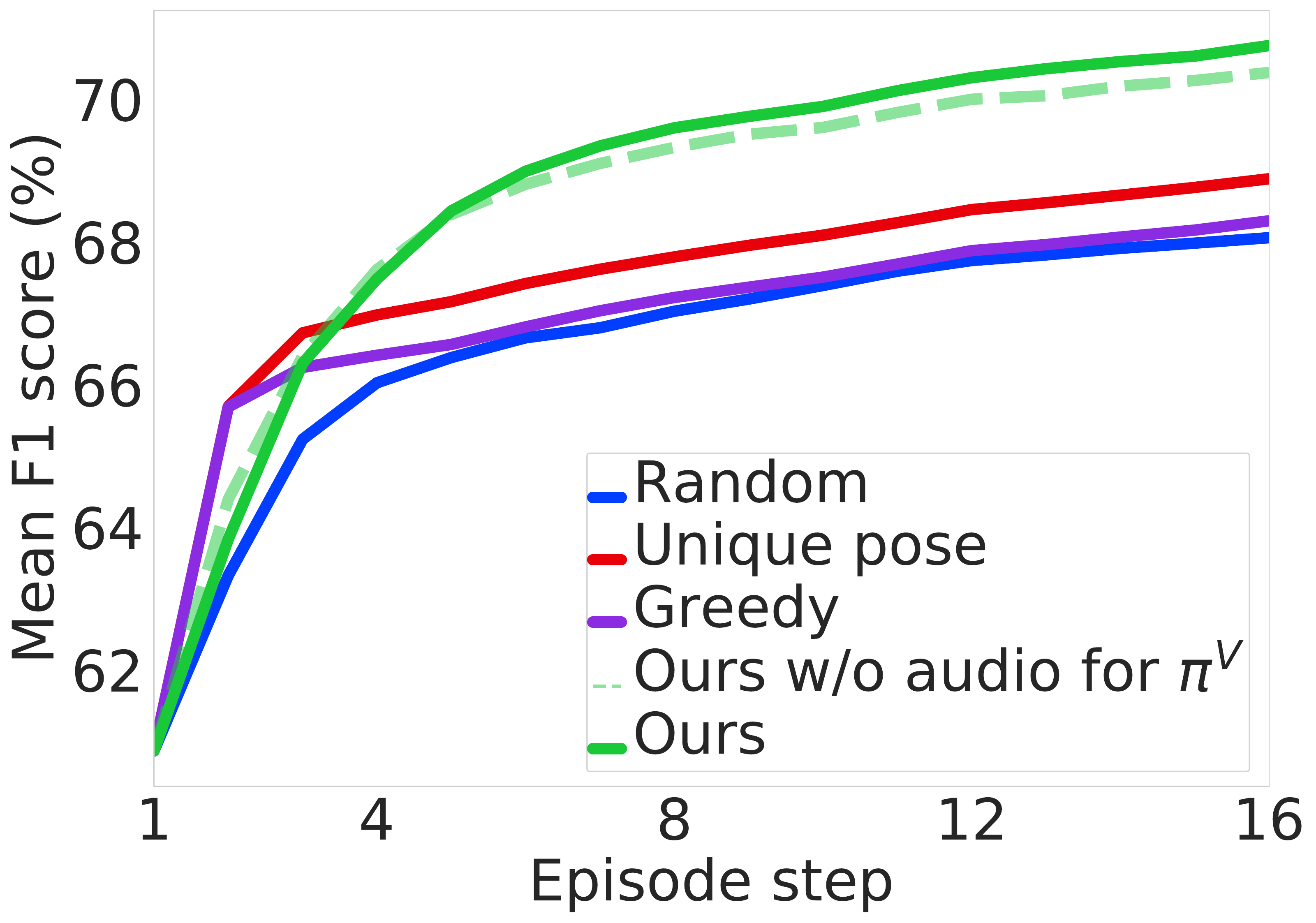}
\caption{
\SM{Effect of ambient environment sounds on \emph{active mapping}
}}
    \label{fig:am_amb}
    \vspace{-0.3cm}
\end{figure}

We also test our model’s robustness to ambient and background sounds by inserting a non-speech sound (\eg running AC, dog barking, etc.)
at a random location outside the egos’ trajectories.
Although quite
challenging, 
our model performs better than 
the baselines for both \emph{passive} (Table~\ref{table:pm_amb})
and \emph{active mapping}
(Fig.~\ref{fig:am_amb}).
Hence, even without explicit audio separation, our model is able to implicitly ground its audio representations in the corresponding pose features for accurate mapping.

}

\subsection{Unheard sounds}\label{sec:supp_unheard}

\begin{table}[t]
  \centering
    \scalebox{0.82}{
    \setlength{\tabcolsep}{2pt}
    \begin{tabular}{lcc}
    \toprule
    Model           & {F1 score $\uparrow$} & {IoU $\uparrow$} \\    \midrule
    All-occupied     &  63.4 &  48.8\\
    Register-inputs      & 72.6 &  60.1\\   
    OccAnt~\citep{ramakrishnan2020occupancy}   & 74.5 & 62.7 \\   
    AV-Floorplan~\citep{purushwalkam2021audio}  & 79.0 & 67.7 \\   
    \textbf{Ours}                       &  \textbf{81.6} & \textbf{71.1} \\
    \midrule
    Ours w/o vision    & 72.6  & 60.1 \\ 
    Ours w/o audio     & 78.1 & 66.7 \\    
    Ours w/o $E^{'}_{i}$'s speech      & 81.3 & 70.7 \\    
    Ours w/o shared mapping      & 80.7 & 70.0 \\
    \bottomrule
  \end{tabular}
  }
  \caption{
  \emph{Passive mapping} performance ($\%$) on \emph{unheard} sounds. 
  } 
  \label{table:pm_unhrd}
\end{table}

\begin{figure}[t] 
    \centering
    \includegraphics[width=0.8\linewidth]{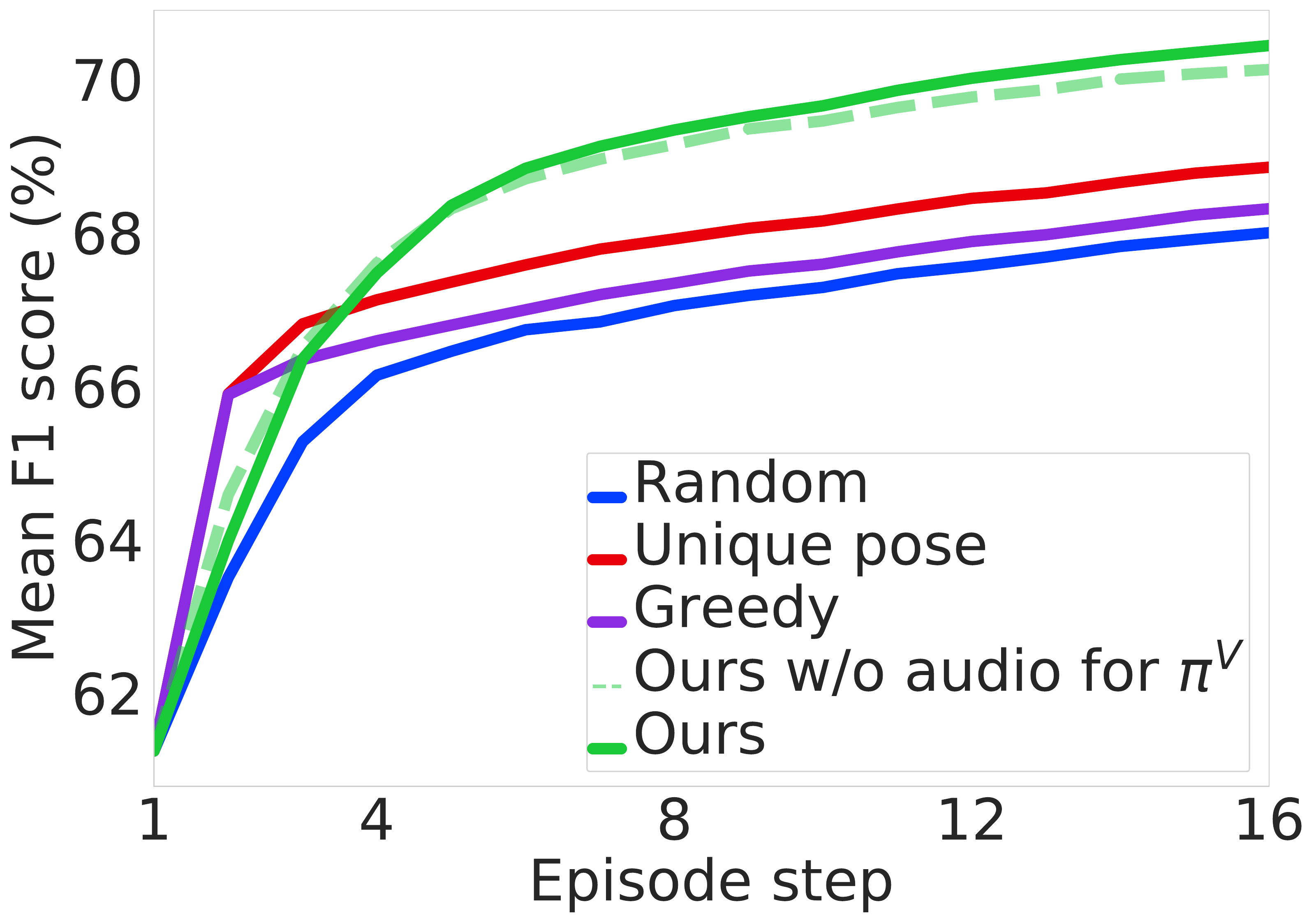}
    \caption{
    \emph{Active mapping} performance vs. episode step on \emph{unheard} sounds. 
    }
\label{fig:am_unhrd}
\end{figure}

In Sec.~\ref{sec:map_pred_results}, we showed results with \emph{heard} sounds (Sec.~\ref{sec:experiments}), \ie the \emph{anechoic} speech sounds uttered by the egos are shared between train and test splits. However, due to our use of \emph{unseen} environments in test (Sec.~\ref{sec:experiments}), the spatial speech sounds input to our model during test are not heard in training. To make the evaluation even more challenging, we conduct a parallel experiment here, where even the anechoic speech is distinct from what's used in training, which we call as the \emph{unheard} sound setting (Sec.~\ref{sec:experiments}). 

Table~\ref{table:pm_unhrd} shows our \emph{passive mapping} results in the \emph{unheard} sound setting.
Our model is able to retain its performance margins over all baselines even in this more challenging scenario. 

We notice a similar trend upon evaluating our model for \emph{active mapping} on \emph{unheard} sounds. Fig.~\ref{fig:am_unhrd} shows that our model is able to generalize to novel sounds better than all baselines. 

This indicates that both our mapper $f^M$ and visual sampling policy $\pi^V$ are able to learn useful spatial cues from audio that are agnostic of the speech content and semantics.

\subsection{Visual budget value}\label{sec:supp_B_value}
\begin{figure}[t]
    \centering
    \begin{subfigure}[b]{0.49\linewidth}
    \centering
    \includegraphics[width=\linewidth]{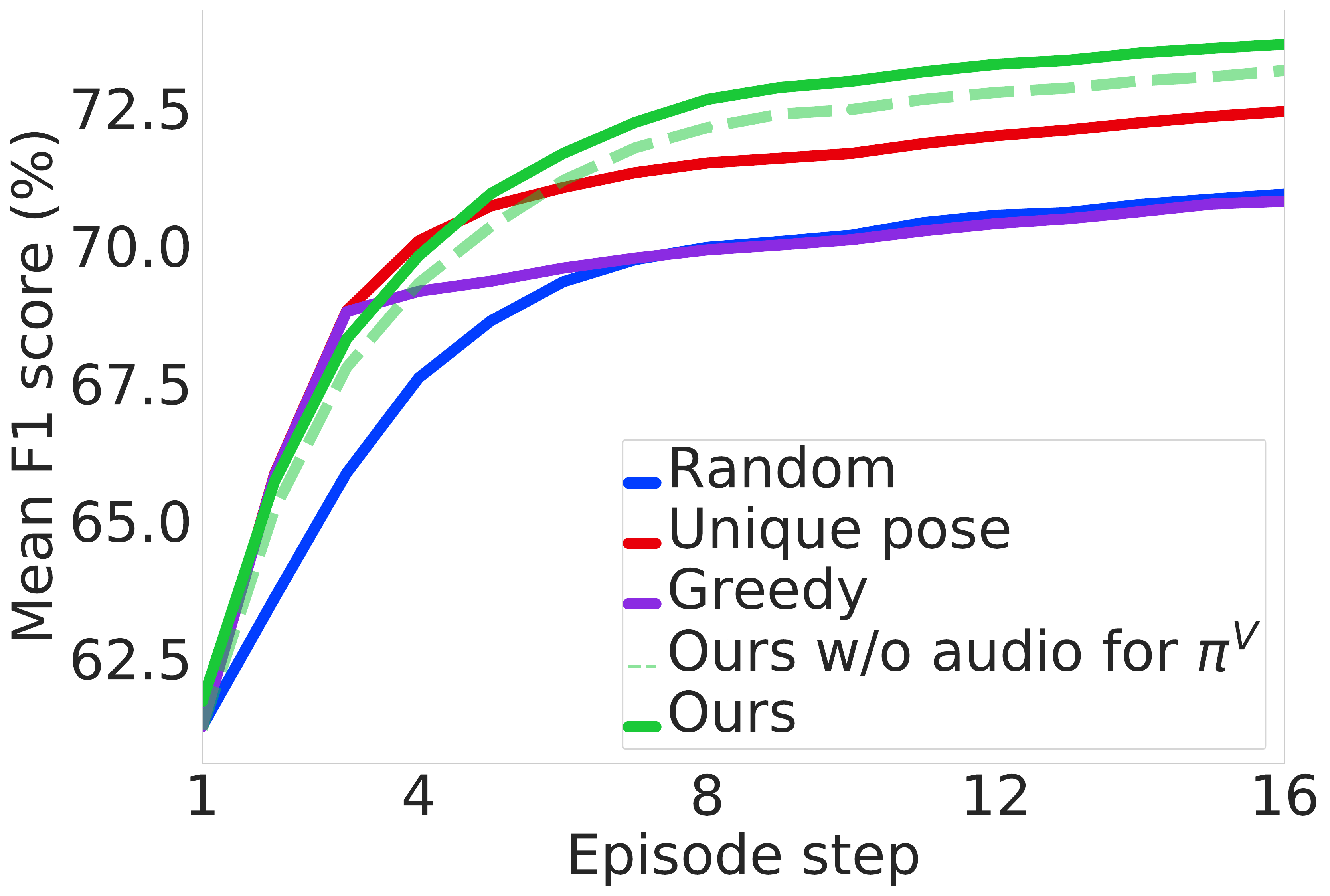}
    \caption{$B = 4$}
    \label{fig:am_tgt64_b4}
    \end{subfigure}\hfill
    \begin{subfigure}[b]{0.49\linewidth}
    \centering
    \includegraphics[width=\linewidth]{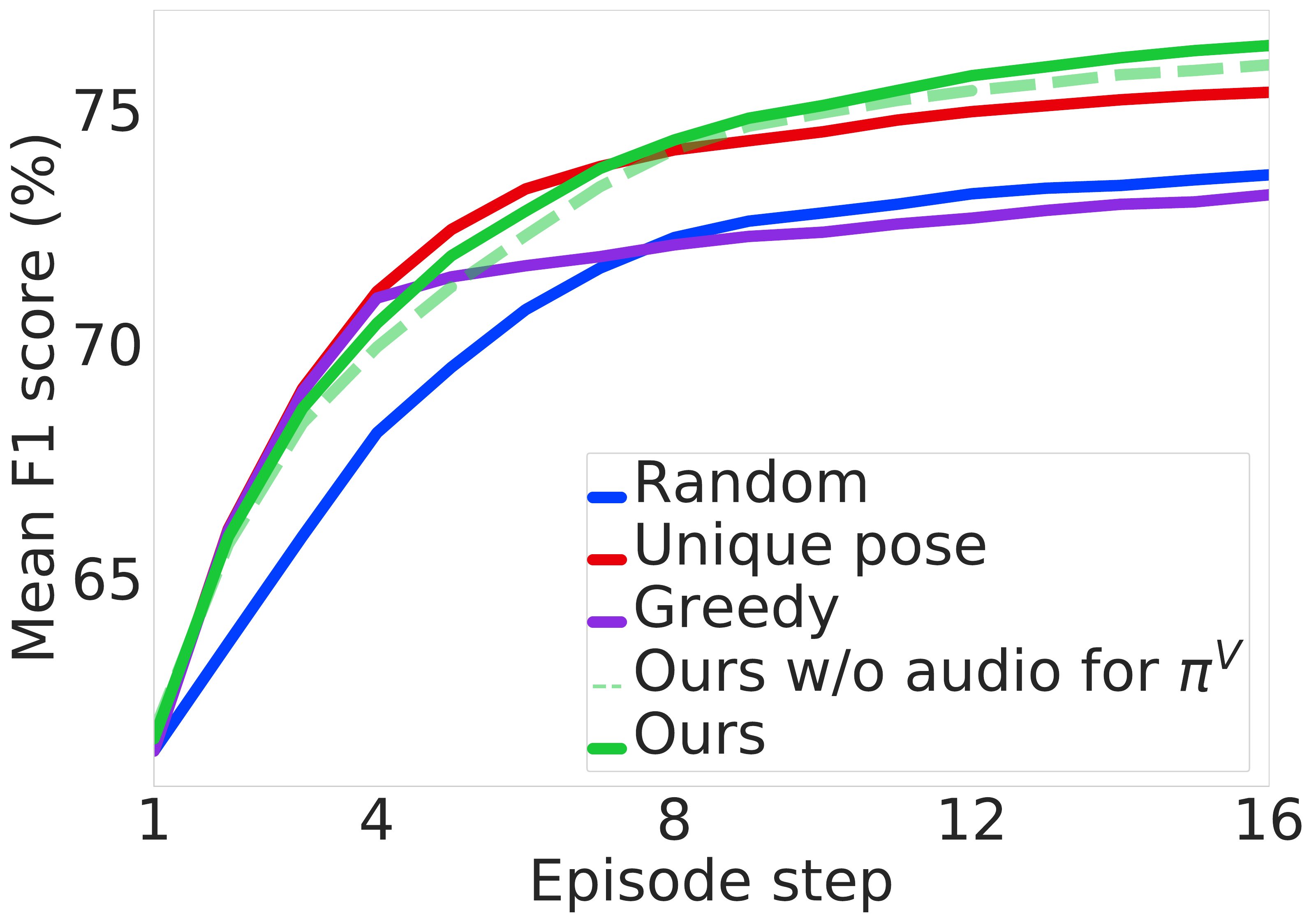}
    \caption{$B = 6$}
    \label{fig:am_tgt64_b6}
    \end{subfigure}\hfill
\caption{
\emph{Active mapping} performance vs. episode step with $B \in \big\{4, 6\big\}$.
}
\label{fig:am_b4_b6}
\end{figure}

So far, we have shown \emph{active mapping} results with the visual budget set to $B=2$ (Sec.~\ref{sec:map_pred_results} and Fig.~\ref{sec:task}). To analyze the effect of larger values of $B$, we show our \emph{active mapping} performance for $B \in \big\{4, 6\big\}$ in Fig.~\ref{fig:am_b4_b6}. Our model outperforms all baselines even for these larger $B$ values. We also observe that the lower the visual budget, the higher the performance margins are for our model. This shows that our model is particularly more robust to the lack of visuals in extremely low-resource settings.

\subsection{Sensor noise}\label{sec:supp_sensor_noise}
\begin{table}[t]
  \centering
    \scalebox{0.82}{
    \setlength{\tabcolsep}{2pt}
    \begin{tabular}{lcc}
    \toprule
    Model           & {F1 score $\uparrow$} & {IoU $\uparrow$}\\    \midrule
    All-occupied     & 63.0 & 48.3  \\
    Register-inputs      & 72.3 & 59.7\\   
    OccAnt~\citep{ramakrishnan2020occupancy}   & 74.7& 63.0 \\   
    AV-Floorplan~\citep{purushwalkam2021audio}  & 77.6 & 65.8\\   
    \textbf{Ours}                       &  \textbf{79.1} & \textbf{68.0} \\
    \midrule
    Ours w/o vision    & 72.6  & 60.0  \\ 
    Ours w/o audio     & 76.7  & 65.1  \\    
    Ours w/o $E^{'}_{i}$'s speech      & 78.8 & 67.7  \\    
    Ours w/o shared mapping      & 78.5 & 67.2  \\
    \bottomrule
  \end{tabular}
  }
  \caption{
  \emph{Passive mapping} performance ($\%$) with sensor noise. 
  } 
  \label{table:pm_noise}
\end{table}

\begin{figure}[t] 
    \centering
    \includegraphics[width=0.8\linewidth]{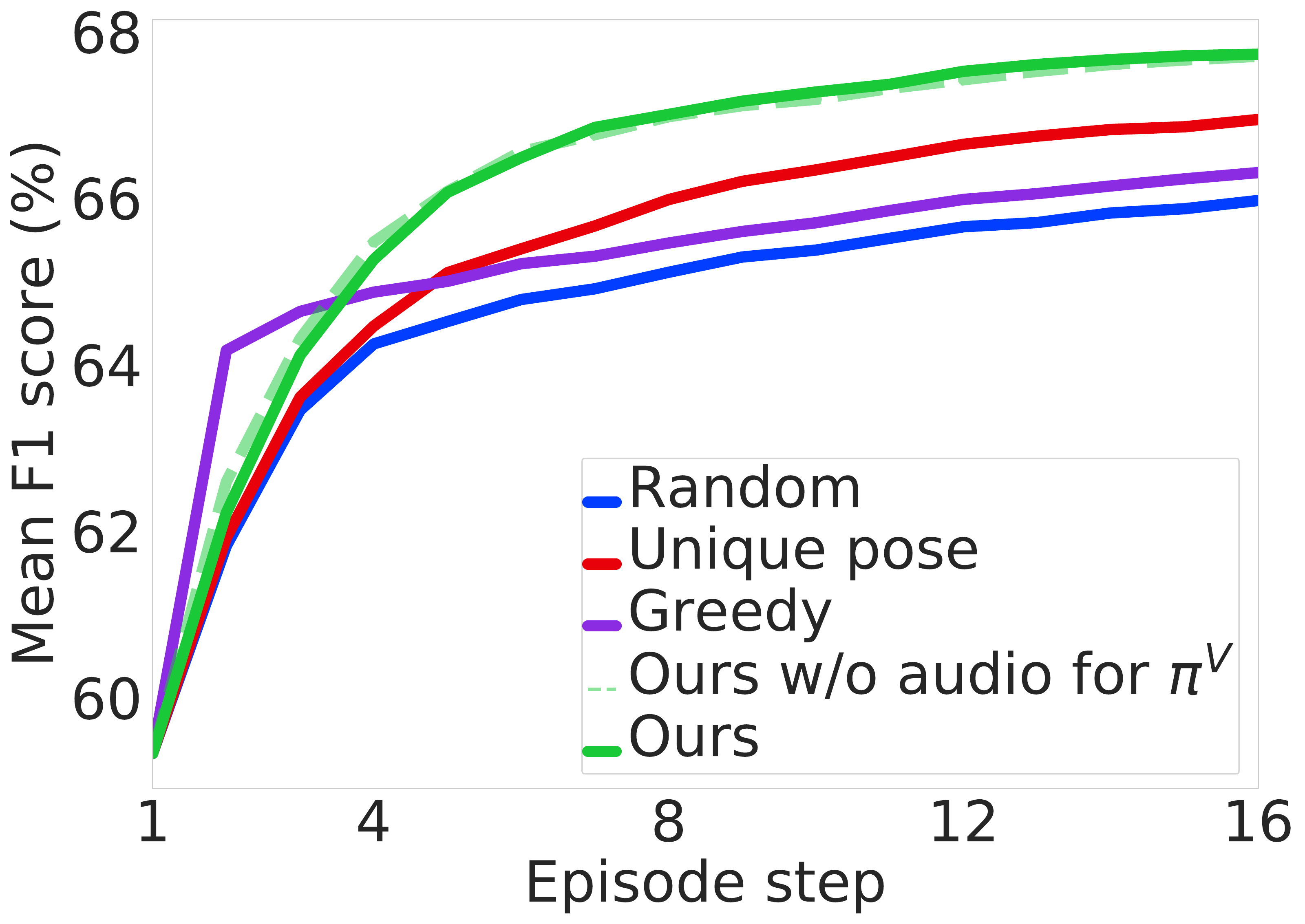}
    \caption{
    \emph{Active mapping} performance vs. episode step with sensor noise. 
    }
\label{fig:am_noise}
\end{figure}

Here, we test our model's robustness to sensor noise by adding noise of the appropriate type individually to each sensor. For RGB images, we sample the noise from a Gaussian distribution with a mean of $0$ and a standard deviation of $0.2$~\citep{ramakrishnan2020occupancy, savva2019habitat}. For depth, we use the Redwood depth noise model~\citep{7299195, ramakrishnan2020occupancy, savva2019habitat}, where the amount of noise is higher for higher depth values and vice-versa. Following ~\citep{ramakrishnan2020occupancy}, we sample pose noise from a truncated Gaussian with a mean of $0.025$ m and a standard deviation of $0.001$ m for the spatial location component of an ego pose $\big((x,y) \text{ in Sec.~\ref{sec:task}}\big)$. For orientation $\theta$ (Sec.~\ref{sec:task}), we use another truncated Gaussian with a mean of $0.9^\circ$ and a standard deviation of $0.057^\circ$. Both distributions are truncated at 2 standard deviations. For our multi-channel microphones (Sec.~\ref{sec:task}), we add a high amount of noise (SNR of 40 dB)~\citep{chen2020soundspaces} using a standard noise model~\citep{7299195, 8461723}. 

Table~\ref{table:pm_noise} and Fig.~\ref{fig:am_noise} report our \emph{passive} and \emph{active mapping} performance, respectively, in the face of sensor noise. In both settings, although our model's performance declines in comparison to the noise-free setting (cf. Table~\ref{table:pm_main} and Fig.~\ref{fig:am_b2}), it generalizes better than all baselines, thereby underlining the effectiveness of our method.

\subsection{Target map size}\label{sec:supp_target_map_size}

\begin{table}[t]
  \centering
    \scalebox{0.82}{
    \setlength{\tabcolsep}{2pt}
    \begin{tabular}{lcc|cc}
    \toprule
                    &   \multicolumn{2}{c|}{$H = W = 8 \text{ m}$} &  \multicolumn{2}{c}{$H = W =  9.6 \text{ m}$}\\
    Model           & {F1 score $\uparrow$} & {IoU $\uparrow$} & {F1 score $\uparrow$} & {IoU $\uparrow$}\\    \midrule
    All-occupied     & 53.5 & 37.9  & 46.4 & 31.2\\
    Register-inputs      & 65.9 & 53.4 & 61.6 & 49.6\\   
    OccAnt~\citep{ramakrishnan2020occupancy}   & 67.8 & 55.7 & 63.0 & 51.3\\   
    AV-Floorplan~\citep{purushwalkam2021audio}  & 71.4 & 59.1 & 68.7 & 53.1\\   
    \textbf{Ours}                       &  \textbf{73.4} & \textbf{60.7} & \textbf{72.0} & 54.4\\
    \midrule
    Ours w/o vision    &  66.1 & 53.5 & 62.6 & 50.3\\ 
    Ours w/o audio     & 71.1 & 58.1 & 63.8 & 51.3\\    
    Ours w/o $E^{'}_{i}$'s speech      & 73.3 & 60.5 & 67.6 & 54.0\\    
    Ours w/o shared mapping      & 72.9 & 60.3 & 68.0 & \textbf{54.5}\\
    \bottomrule
  \end{tabular}
  }
  \caption{
  \emph{Passive mapping} performance ($\%$) for larger target map sizes. 
  } 
  \label{table:pm_tgt_mp_sz}
\end{table}

\begin{figure}[t]
    \centering
    \begin{subfigure}[b]{0.49\linewidth}
    \centering
    \includegraphics[width=\linewidth]{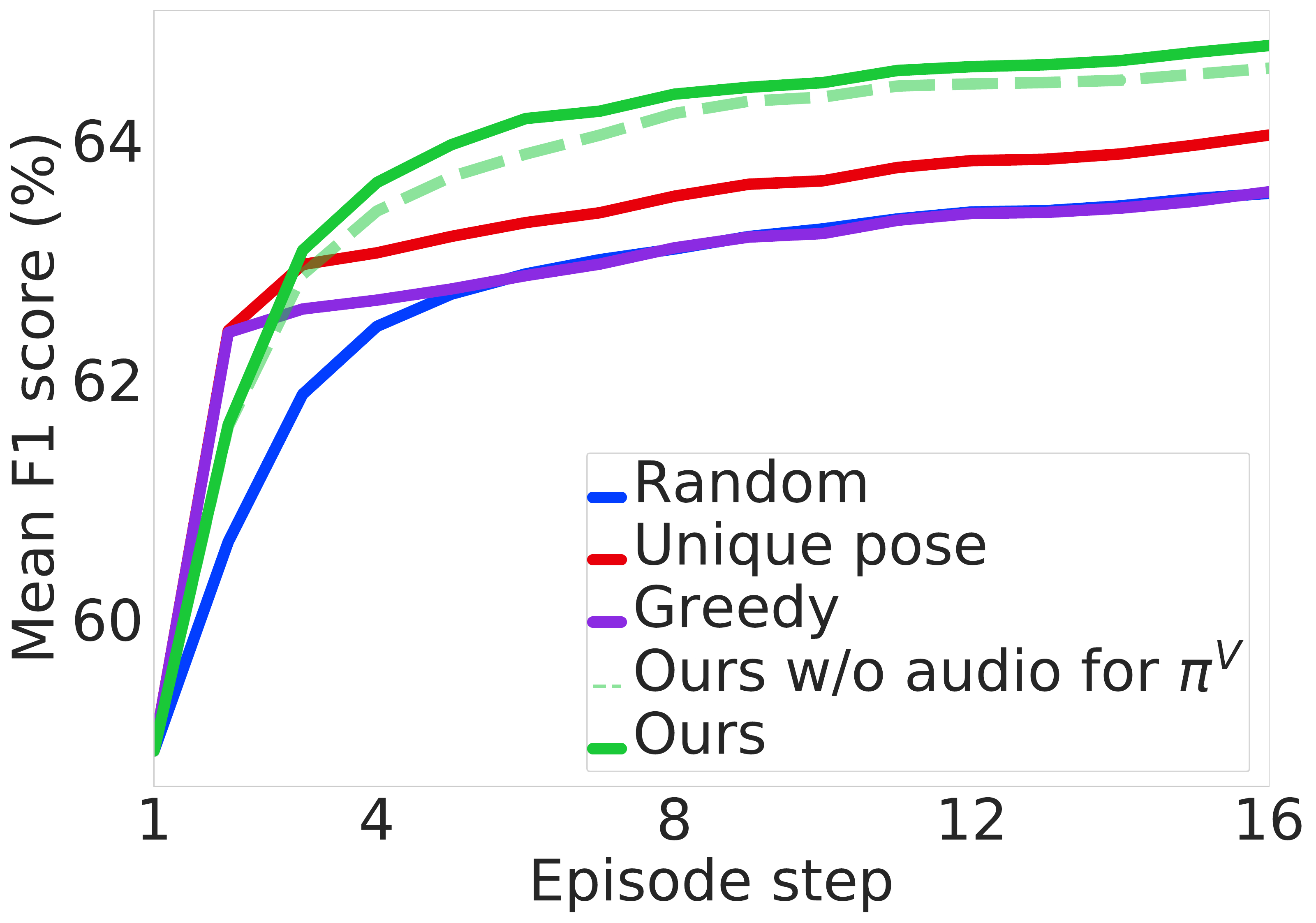}
    \caption{$H = W = 8 \text{ m}$}
    \label{fig:am_tgt80_b2}
    \end{subfigure}\hfill
    \begin{subfigure}[b]{0.49\linewidth}
    \centering
    \includegraphics[width=\linewidth]{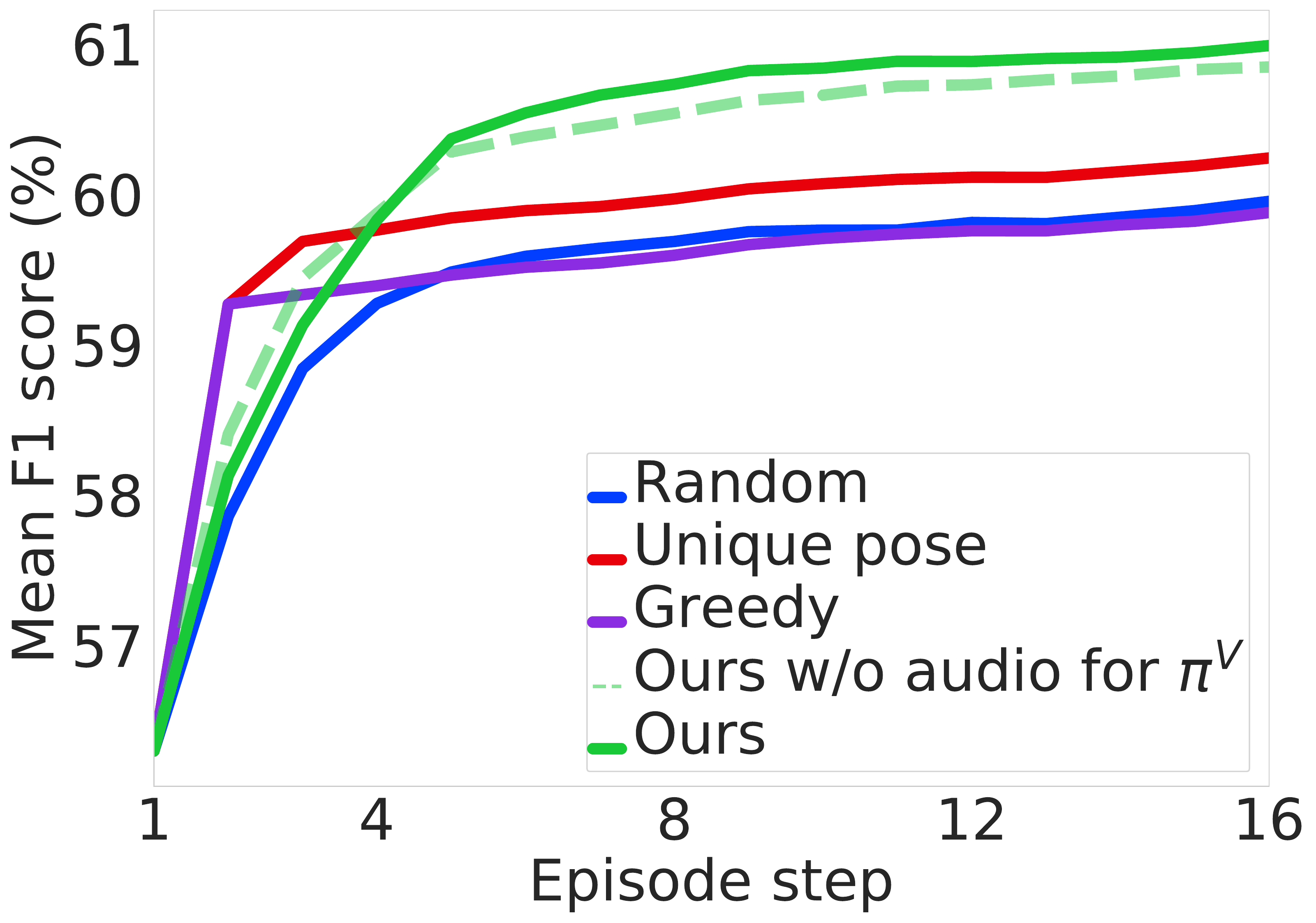}
    \caption{$H = W = 9.6 \text{ m}$}

    \label{fig:am_tgt96_b2}
    \end{subfigure}\hfill
\caption{
\emph{Active mapping} performance vs. episode step for larger target map sizes.
}
\label{fig:am_tgt_mp_sz}
\end{figure}

In Sec.~\ref{sec:map_pred_results}, we showed mapping results with $H \times W = 6.4 \times 6.4 \text{ m}^2 (\sim 41 \text{ m}^2)$, where $H$ and $W$ denote the height and width of the ground-truth local $360^\circ$ FoV maps (Sec.~\ref{sec:mapper}). 
To analyze the impact of larger target map sizes on the mapping quality, we also test our model with $H \times W \in \big\{8 \times 8 \text{ m}^2 (64 \text{ m}^2), 9.6 \times 9.6 \text{ m}^2 (\sim 92 \text{ m}^2)\big\}$. Table~\ref{table:pm_tgt_mp_sz} and Fig.~\ref{fig:am_tgt_mp_sz} show the corresponding results for \emph{passive} and \emph{active mapping}, respectively.  In both cases, our model outperforms all baselines by a substantial margin, showing that our method is also robust to higher target map sizes.

\SMC{\subsection{Different ego initializations}}\label{sec:supp_ego_inits}

\begin{table}[t]
  \centering
    \scalebox{0.82}{
    \setlength{\tabcolsep}{2pt}
    \begin{tabular}{lcc|cc}
    \toprule
                    &   \multicolumn{2}{c|}{Facing away} &  \multicolumn{2}{c}{Separated by occlusion}\\
    Model           & {F1 score $\uparrow$} & {IoU $\uparrow$} & {F1 score $\uparrow$} & {IoU $\uparrow$}\\    \midrule
    OccAnt~\citep{ramakrishnan2020occupancy}   & 75.2 & 63.7 & 75.3 & 62.8 \\   
    AV-Floorplan~\citep{purushwalkam2021audio}  & 79.6 & 68.7 & 80.1 & 69.0\\   
    \textbf{Ours}                       &  \textbf{82.9} & \textbf{72.8} & \textbf{83.0} & \textbf{71.8}\\
    \bottomrule
  \end{tabular}
  }
  \caption{
  \emph{Passive mapping} performance ($\%$) for different ego initializations. 
  } 
  \label{table:pm_diff_ego_inits}
\end{table}

\begin{figure}[t]
    \centering
    \begin{subfigure}[b]{0.49\linewidth}
    \centering
    \includegraphics[width=\linewidth]{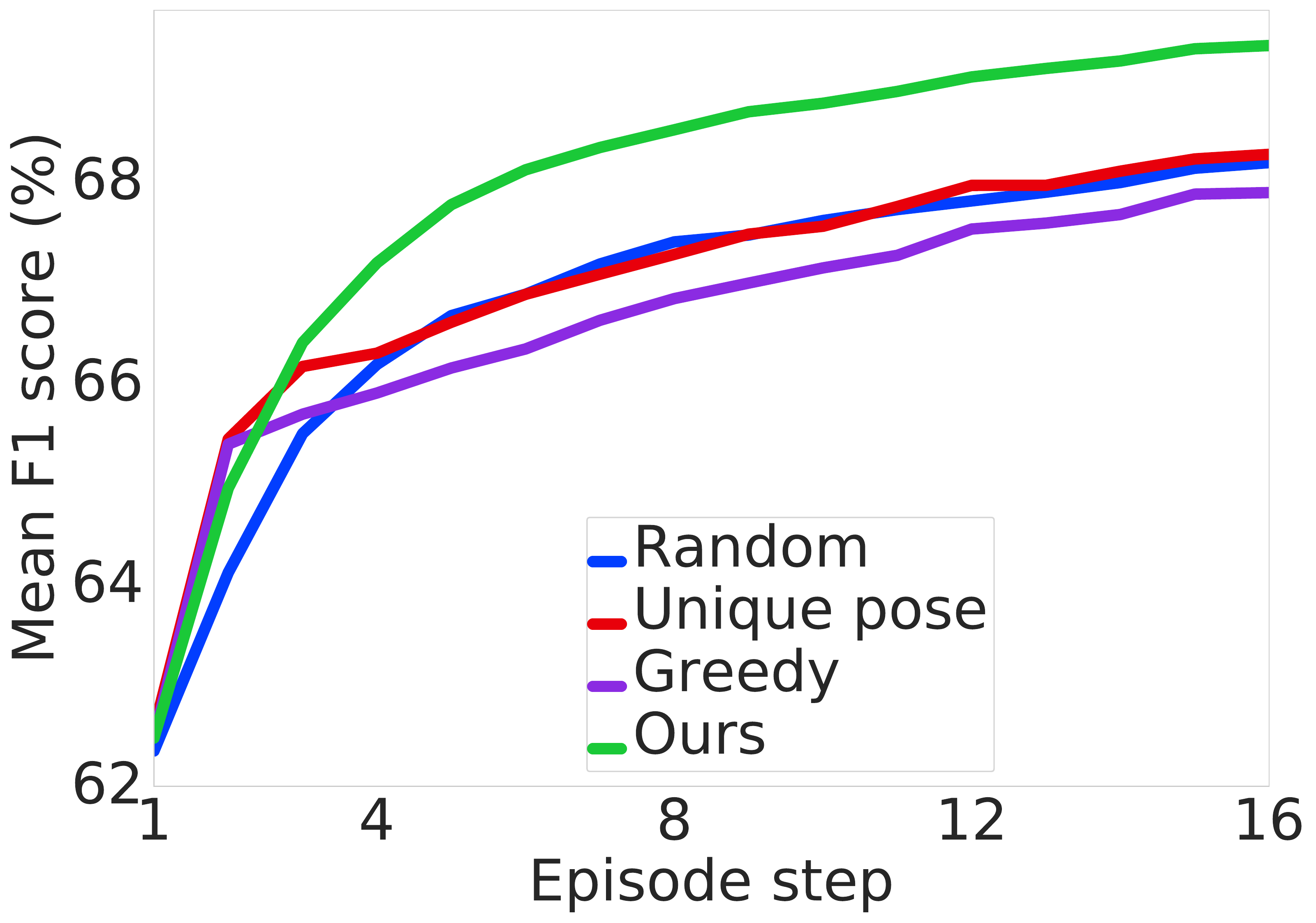}
    \caption{Facing away}
    \label{fig:am_agntsFcAwy}
    \end{subfigure}\hfill
    \begin{subfigure}[b]{0.49\linewidth}
    \centering
    \includegraphics[width=\linewidth]{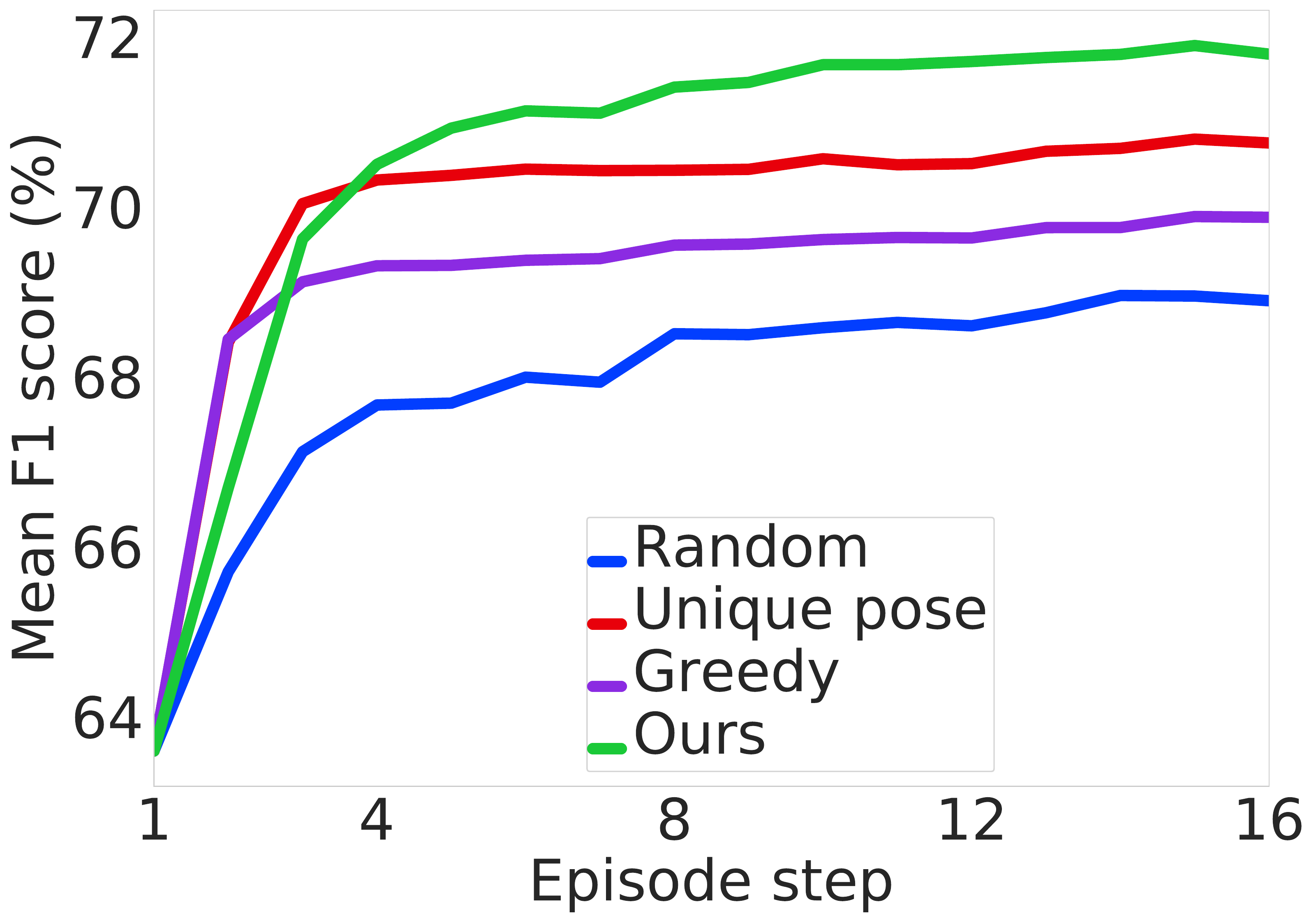}
    \caption{Separated by occlusion}

    \label{fig:am_agntsOcc}
    \end{subfigure}\hfill
\caption{
\emph{Active mapping} performance for different ego initializations.
}
\label{fig:am_diff_ego_inits}
\end{figure}

\SMC{Here, we study the effect of ego initialization by considering two cases: 1) the egos initially face away from each other, and 2) the egos are initially separated by an occlusion. In both cases, our model outperforms all baselines on both passive (Table~\ref{table:pm_diff_ego_inits}) and active mapping (Fig.~\ref{fig:am_diff_ego_inits}), showing that our model is robust to different ego initializations.
}

\SMC{\subsection{Multiple random dataset splits}}\label{sec:supp_mult_data_splits}

\begin{table}[t]
  \centering
    \scalebox{0.82}{
    \setlength{\tabcolsep}{2pt}
    \begin{tabular}{lcc}
    \toprule
    Model           & {F1 score $\uparrow$} & {IoU $\uparrow$}\\    \midrule
    OccAnt~\citep{ramakrishnan2020occupancy}   & 74.4 & 63 \\   
    AV-Floorplan~\citep{purushwalkam2021audio}  & 78.8 & 67.6 \\   
    \textbf{Ours}                       & \textbf{81} & \textbf{70.6} \\
    \bottomrule
  \end{tabular}
  }
  \caption{
  Average \emph{passive mapping} performance ($\%$) over 3 random data splits. 
  } 
  \label{table:pm_mult_data_splits}
\end{table}

\begin{figure}[t] 
    \centering
    \includegraphics[width=0.8\linewidth]{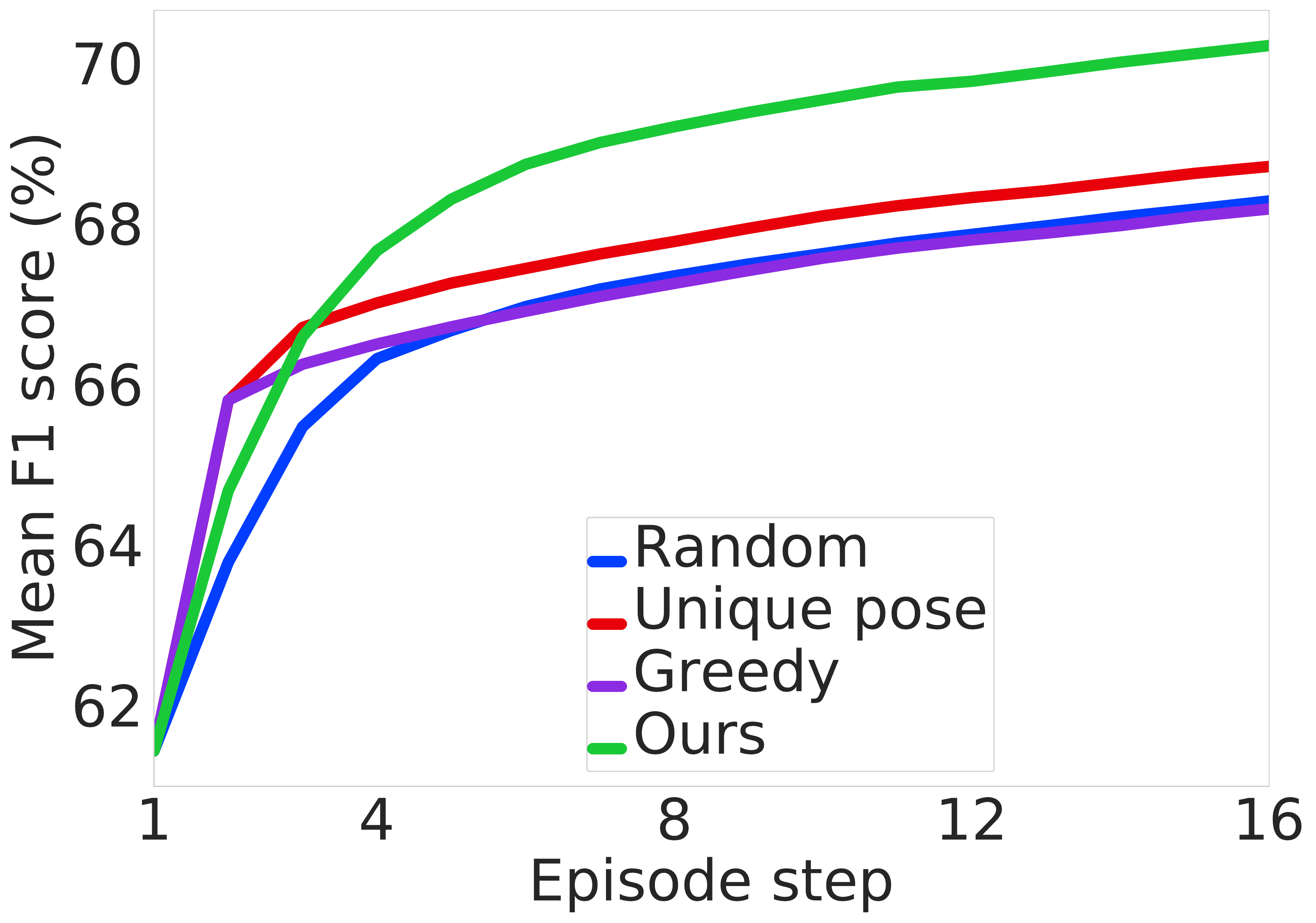}
    \caption{
    Average \emph{active mapping} performance over 3 random data splits vs. episode step. 
    }
\label{fig:am_mult_data_splits}
\end{figure}

\SMC{
Here, we gauge the effect of multiple random dataset splits on our model performance. In each random split, we use the same distribution of train/val/test scenes and episode counts as mentioned in Sec.~\ref{sec:experiments}, but instantiate each episode with a new seed. Table~\ref{table:pm_mult_data_splits} and Fig.~\ref{fig:am_mult_data_splits} report the passive and active mapping performance averaged over 3 random splits, respectively. We observe that our model is more robust to different dataset instantiations than all baselines across both passive and active mapping.
}

\SMC{\subsection{Training data size}}\label{sec:supp_train_data_size}
\begin{figure}[t] 
    \centering
    \includegraphics[width=0.8\linewidth]{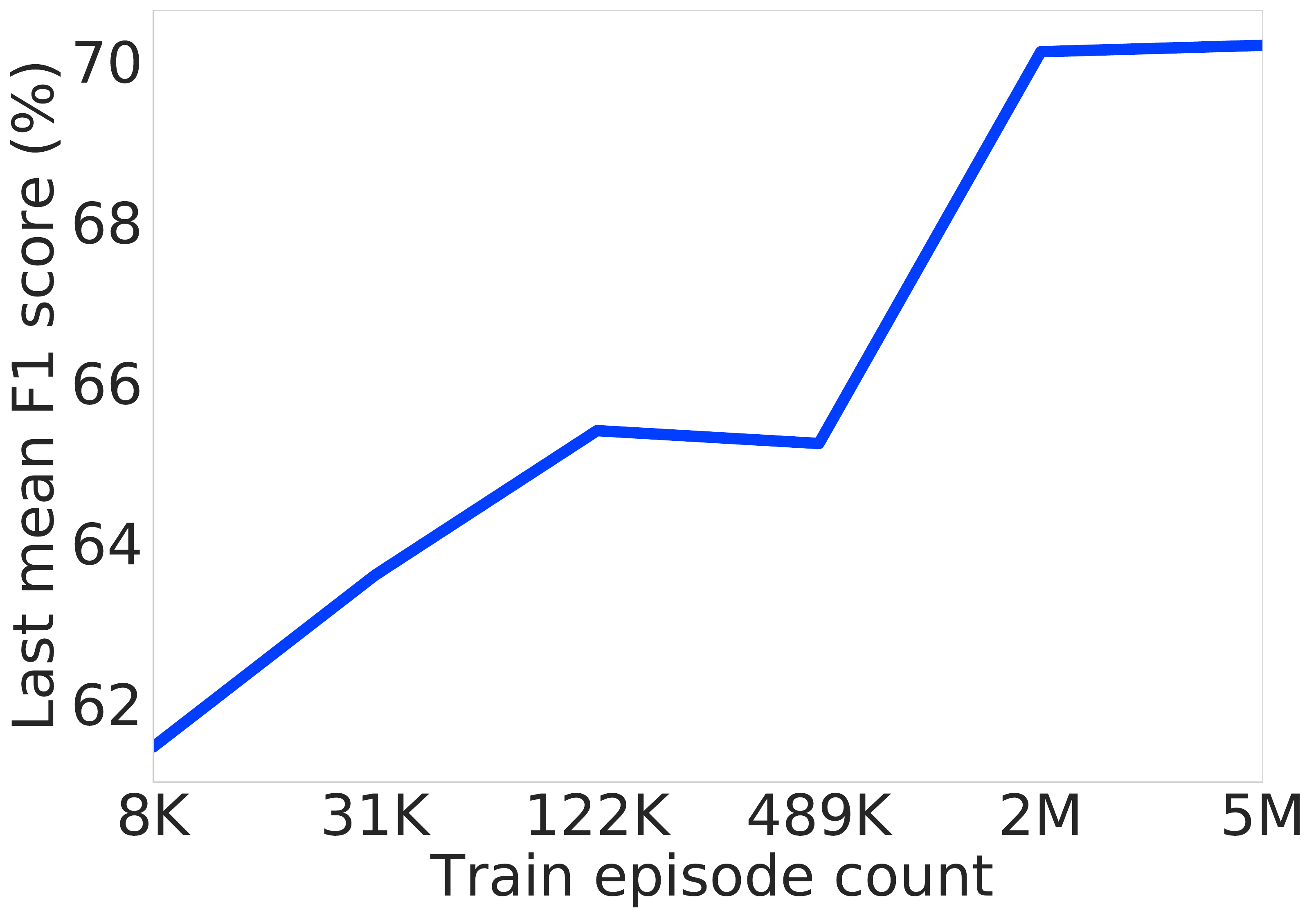}
    \caption{
    \emph{Active mapping} performance on \emph{unheard} sounds vs. training data size. 
    }
\label{fig:am_train_data_size}
\end{figure}

\SMC{
Here, we analyze the effect of the training data size on the active mapping quality. Fig.~\ref{fig:am_train_data_size} shows the final active mapping F1 score as a function of the number of training episodes. The mapping performance significantly improves with more training data up to 2 million episodes and then flattens between 2 and 5 million episodes.
}

\subsection{Dataset details}\label{sec:supp_dataset}
Here, we provide additional dataset details. 

\paragraph{Visual data.}
All RGB-D images in our experiments  have a resolution of $128 \times 128$. 

To generate the topdown occupancy maps, we threshold the local pointcloud computed from the $90^\circ$ FoV depth images (Sec.~\ref{sec:inp_prep}) using a lower and upper height limit of 0.2 and 1.5 m, respectively, such that a map cell is considered occupied if there is a 3D point for it in the 0.2-1.5 m range, and free otherwise.

To generate an estimate of the scene map, we register the estimates of 
ground-truth local $360^\circ$ FoV
maps, $\tilde{M}_{i, j}$s onto a shared scene map $\tilde{M}$ (Sec.~\ref{sec:mapper}) and maintain a count of the number of updates undergone by every cell in the shared map. To register a local estimate  $\tilde{M}_{i, j}$, we first translate and rotate $\tilde{M}_{i, j}$ within $\tilde{M}$  on the basis of its normalized pose $P_{i, j}$. Next, we add $\tilde{M}_{i, j}$ with the corresponding part of $\tilde{M}$ and update the counter for every map cell that's been changed through the registration. We repeat this process for every $\tilde{M}_{i, j}$ in the episode. Finally, we normalize the updated $\tilde{M}$ by dividing each cell in it by its number of updates from the counter, and thresholding at 0.5. In our experiments, $\tilde{M}$ covers a maximum area of $128.4 \times 128.4 \text{ m}^2$.

\paragraph{Audio data.}
For each conversation episode, we randomly choose $2$ speakers from the same split -- \emph{heard} or \emph{unheard} (Sec.~\ref{sec:experiments}). Starting at a random time in the audio clip for each speaker, we choose contiguous 3s slices from each clip for $T$ steps to use as the anechoic audio for the two egos in the episode, where $T$ denotes the episode length (Sec.~\ref{sec:task}). Further, we normalize each
slice to have the same RMS value of 400 across the whole
dataset, where all audio is sampled at 16 kHz and stored
using the standard 16-bit integer format. 

To generate the spectrograms, we convolve a speech slice with the appropriate 9-channel RIR sampled at 16 kHz and compute its STFT with a Hann window of 31.93 ms, hop length of 8.31 ms, and FFT size of 511 to generate 9-channel magnitude spectrograms, where each channel has 256 frequency bins and 257 overlapping temporal windows. We assume access to detected and separated speech from the egos at all times since on-device microphones of AR glasses can tackle  nearby and distant speaker detection~\citep{Jiang2022EgocentricDM} and separation~\citep{Patterson2022DistanceBasedSS}.

\subsection{Baselines}\label{sec:supp_baselines}
Here, we provide additional implementation details for our \emph{active mapping} baselines for reproducibility (Sec.~\ref{sec:experiments}).

\begin{itemize}
    \item \textbf{Random.} At each step $t$, we generate a random number between 0 and 1 from a uniform distribution. Depending on which quartile of the 0-1 range the random number lies in, we skip visual frames for both egos, sample for just one ego, or sample for both egos.  
    \item \textbf{Greedy.} Starting at $t=2$, we sample visual frames for both egos at every step until we run out of the visual budget $B$. If the value of $B$ is such that it allows sampling only one visual frame at a certain step (\ie $B$ is odd), we randomly choose the ego for which we sample the frame at that step.
    \item \textbf{Unique-pose.} To implement this baseline, we keep track of the egos' poses during an episode. At any step $t$, we sample the frame for an ego if it's current pose has never been assumed before by either of the egos in that episode.  
\end{itemize}

\subsection{Architecture and training}\label{sec:supp_arc_n_train}
Here, we provide our architecture and additional training details for reproducibility. We will release our code. 

\subsubsection{Policy architecture}~\label{sec:pol_arc}
\paragraph{Visual encoder.} To encode local occupancy map inputs, our policy $\pi^V$ (Sec.~\ref{sec:policy}) uses a 6-layer CNN consisting of 5 convolutional (conv.) layers followed by an adaptive average pooling layer. The first three conv. layers use a kernel size of 4 and a stride of 2, while the last two conv. layers use a kernel size of 3 and a stride of 1. All conv. layers use a zero padding of 1, except for the third conv. layer, which uses a zero padding of 2. The number of output channels of the conv. layers are [64, 64, 128, 256, 512], respectively. Each convolution is followed by a leaky ReLU~\citep{ICML-2010-NairH, Sun2015DeeplyLF} activation with a negative slope of 0.2, and a Batch Normalization~\citep{pmlr-v37-ioffe15} of $1e^{-5}$. The adaptive average pooling layer reduces the output of the last conv. layer to a feature of size $1\times1\times512$. 

To encode RGB images (Sec.~\ref{sec:policy}), $\pi^V$ uses a separate CNN with 5 conv. layers and an adaptive average pooling layer. Each conv. layer has a kernel size of 4, stride of 2 and zero padding of 1. The number of output channels are [64, 64, 128, 256, 512], respectively. Similar to the occupancy map encoder, each convolution is followed by a leaky ReLU~\citep{ICML-2010-NairH, Sun2015DeeplyLF} activation with a negative slope of 0.2 and a Batch Normalization~\citep{pmlr-v37-ioffe15} of $1e^{-5}$, and the adaptive average pooling layer reduces the output of the last conv. layer to a feature of size $1\times1\times512$. 

We fuse the occupancy and RGB features by concatenating them and passing through a single linear layer that produces a 512-dimensional visual embedding $v$ (Sec.~\ref{sec:policy}).

\paragraph{Speech encoder.}
The speech encoder (Sec.~\ref{sec:policy}) in $\pi^V$ is a CNN with 5 conv. layers and an adaptive average pooling layer. Each conv. layer has a kernel size of 4, stride of 2 and a padding of 1, except for the second conv. layer, which has a kernel size of 8, stride of 4 and padding of 3. The number of channels in the CNN are [64, 64, 128, 256, 512], respectively. Similar to the visual encoder, each conv. layer is followed by a leaky ReLU~\citep{ICML-2010-NairH, Sun2015DeeplyLF} with a negative slope of 0.2 and a Batch Normalization~\citep{pmlr-v37-ioffe15} of $1e^{-5}$. The adaptive average pooling layer reduces the output of the last conv. layer to a feature of size $1 \times 1 \times 512$.

\paragraph{Pose encoder.}
The pose encoder (Sec.~\ref{sec:policy}) in $\pi^V$ is a single linear layer that takes a normalized pose $P$ (Sec.~\ref{sec:inp_prep}) as input and produces a 32-dimensional pose embedding.

\paragraph{Fusion layers.}
We perform linear fusion of  the visual, speech and pose embeddings (Sec.~\ref{sec:policy} and Fig.~\ref{fig:model}) at two levels. The first level has 4 linear layers and the second level has 1 linear layer. Each linear layer produces a 512-dimensional fused feature as its output.

\paragraph{Policy network.} 
The policy network (Sec.~\ref{sec:policy}) comprises a one-layer bidirectional GRU~\citep{NIPS2015_b618c321} with 512 hidden units. The actor and critic networks consist of one linear layer.

\subsubsection{Mapper architecture}
\paragraph{Visual encoder.}
To encode local occupancy map inputs, our shared mapper $f^M$ (Sec.~\ref{sec:mapper}) uses a CNN similar to the one used for encoding occupancy maps in $\pi^V$ (Sec.~\ref{sec:pol_arc}), except that it doesn't have a pooling layer at the end. The RGB encoder (Sec.~\ref{sec:mapper}) in $f^M$ is also similar to the one for $\pi^V$, except that it also doesn't have a pooling layer at the end. We fuse the map and RGB features by concatenating them along the channel dimension, and obtain a $4\times4\times1024$ dimensional feature. 

\paragraph{Speech encoder.}
The speech encoders (Sec.~\ref{sec:mapper}) in $f^M$ are CNNs with 5 layers that share the architecture with the first 5 conv. layers of the speech encoder in $\pi^V$ (Sec.~\ref{sec:pol_arc}), except that the last conv. layer in both encoders has 1024 output channels.

\paragraph{Modality encoder.}
For our modality embedding $\hat{m}$ (Sec.~\ref{sec:mapper}), we maintain a sparse lookup table of 1024-dimensional learnable embeddings, which we index with 0 to retrieve the visual modality embedding ($\hat{m}_V$), 1 to retrieve the modality embedding ($\hat{m}_S$) for the speech from self, and 2 to retrieve the modality embedding ($\hat{m}_{S'}$) for the speech from the other ego. 

\paragraph{Occupancy prediction network.}
The transformer~\citep{vaswani2017attention} (Sec.~\ref{sec:mapper}) in our occupancy prediction network comprises 6 encoder and 6 decoder layers, 8 attention heads, an input and output size of 1024, a hidden size of 2048, and ReLU~\citep{ICML-2010-NairH, Sun2015DeeplyLF} activations. Additionally, we use a dropout~\citep{JMLR:v15:srivastava14a} of 0.1 in our transformer. 

The transpose convolutional network $U$ (Sec.~\ref{sec:mapper}) consists of 6 layers in total. The first 5 layers are transpose convolutions (conv.) layers. The first 4 transpose conv. layers have a kernel size of 4 and stride of 2, and the last transpose conv. layer has a kernel size of 3 and stride of 1. Each transpose conv. has a padding of 1, ReLU~\citep{ICML-2010-NairH, Sun2015DeeplyLF} activation and Batch Normalization~\citep{pmlr-v37-ioffe15}. The number of the output channels for the transpose conv. layers are [512, 256, 128, 64, 2], respectively. The last layer in $U$ is a sigmoid layer (Sec.~\ref{sec:mapper}), which outputs the map estimates. 

\subsubsection{Parameter initialization}
We use the Kaiming-normal~\citep{he2015delving} weight initialization strategy to initialize the weights of all our network modules, except for the pose encoding layers and fusion layers, which are initialized with Kaiming-uniform~\citep{he2015delving} initialization, and the policy network, which is initialized using the orthogonal initialization strategy~\citep{Saxe2014ExactST}. We switch off biases in all network modules, except for the policy network where we set the biases initially to 0. 

\subsubsection{Training hyperparameters.}
\paragraph{Policy training.}
To train our policy $\pi^V$ using DD-PPO~\citep{Wijmans2020DDPPOLN} (Sec.~\ref{sec:training}), we weight the action loss by 1.0, value loss by 0.5, and entropy loss by 0.05. We train our policy on 8 Nvidia Tesla V100 SXM2 GPUs with Adam~\citep{kingma2014adam}, an initial learning rate of $1e^{-4}$ and 8 processes per GPU for 8.064 million policy prediction steps. Among other policy training parameters, we set the clip parameter value to 0.1, number of DD-PPO epochs to 4, number of mini batches to 1, max gradient norm value to 0.5, reward discount factor $\gamma$ to 0.99, and the value of $\lambda$ in the generalized advantage estimation~\citep{Schulman2016HighDimensionalCC} formulation for DD-PPO to 0.95.

\paragraph{Mapper training.}
We train our shared scene mapper $f^M$ (Sec.~\ref{sec:mapper}) with a binary cross entropy loss (Sec.~\ref{sec:training}) on 4 Nvidia Quadro RTX 6000 GPUs until convergence by using Adam~\citep{kingma2014adam}, an initial learning rate of $1e^{-4}$ and a batch size of $24$. 